\documentclass[letterpaper]{article}
\usepackage[preprint]{aaai2027}
\usepackage[hyphens]{url}  
\usepackage{graphicx} 
\usepackage{natbib}  
\usepackage{caption} 
\usepackage{amsmath}
\usepackage{amssymb}
\usepackage{bm}
\usepackage{booktabs}
\usepackage{multirow}
\usepackage{array}
\usepackage{graphicx}
\usepackage{subcaption}
\usepackage[most]{tcolorbox}
\tcbset{
    promptbox/.style={
        enhanced,
        breakable,
        colback=gray!5,
        colframe=gray!60,
        boxrule=0.5pt,
        arc=2pt
    }
}
\usepackage{xcolor}
\usepackage{xcolor}
\usepackage{colortbl}

\definecolor{improve}{RGB}{38,120,80}   
\definecolor{degrade}{RGB}{120,120,120} 

\newcommand{\model}{LLMODE}
\newcommand{\secondbest}[1]{\underline{#1}}
\newcommand{\tbd}{\textbf{TBD}}

\title{LLMODE: Aligning ODEs with LLMs via Gated Token Injection
for Irregular Spatio-Temporal Forecasting}
\author{
Di Zhang,
Jingyang Zhang,
Ziqian Wang,
Chi Zhang,
Yikun Ban,
Ziwei Zhang,
Ruijie Wang\corresponding
}

\affiliations{
Beihang University\\
Beijing, China\\
ruijiew@buaa.edu.cn
}

\begin{document}

\maketitle

Large language models (LLMs) have shown promise for spatio-temporal forecasting, but existing approaches often rely on regularly sampled token sequences and struggle with irregular observations because of temporal asynchrony, representation-space misalignment, and limited context windows. We propose LLMODE, a token-efficient framework for irregular spatio-temporal forecasting with a frozen LLM backbone. LLMODE first uses a graph-aware ODE encoder to reconstruct irregular graph observations as a continuous-time latent trajectory. A Fixed-Budget Perceiver Resampler then compresses this variable-length trajectory into a fixed number of dynamic memory tokens. In parallel, compact statistical descriptors are encoded and resampled into context memory tokens. A dual-source gated cross-attention module injects both memories into the frozen LLM, enabling controlled utilization of external spatio-temporal evidence. Experiments on three real-world urban datasets and two physical-dynamics benchmarks show competitive overall performance, with clearer advantages under sparse or dynamically complex irregular sampling. Additional evaluations on unseen urban regions further demonstrate strong zero-shot generalization without adaptation. 
\section{Introduction}

\begin{figure}[t]
\centering

\captionsetup[subfigure]{skip=-3pt, justification=centering}

\begin{subfigure}[t]{0.95\columnwidth}
    \centering
    \includegraphics[width=\columnwidth,keepaspectratio]{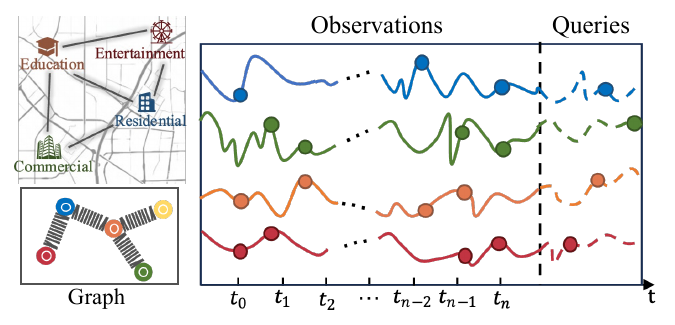}
    \caption{Problem setup}
    \label{fig:1a}
\end{subfigure}

\begin{subfigure}[t]{0.46\columnwidth}
    \centering
    \includegraphics[width=\columnwidth,keepaspectratio]{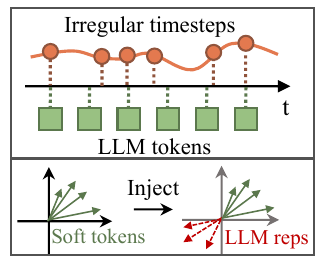}
    \caption{Semantic alignment}
    \label{fig:1b}
\end{subfigure}
\begin{subfigure}[t]{0.46\columnwidth}
    \centering
    \includegraphics[width=\columnwidth,keepaspectratio]{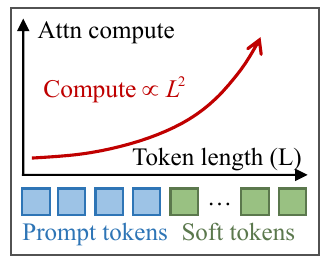}
    \caption{Context-length constraint}
    \label{fig:1c}
\end{subfigure}

\vspace{-1ex}

\caption{
(a) Problem setup for irregular spatio-temporal graph forecasting.
(b) Challenge 1: Semantic alignment between soft tokens and LLM semantic space.
"LLM reps" refer to the representations in the LLM's hidden space.
(c) Challenge 2: Context-length constraint in conditioning LLMs.
}

\label{fig:align}

\end{figure}

Irregularly sampled spatio-temporal data are common in urban sensing and physical systems, as illustrated in Figure~\ref{fig:align}. Traffic flows, crime reports, mobility records, and interacting entities may be observed at non-uniform timestamps, with missing values and asynchronous updates across locations or objects. Because these observations evolve over relational structures, accurate forecasting must jointly model temporal irregularity, incomplete measurements, and graph-structured dependencies.

Large language models (LLMs) have recently been applied to spatio-temporal forecasting through token-based conditioning. UrbanGPT~\cite{UrbanGPT}, for example, compresses spatio-temporal signals into soft tokens for a pretrained LLM. However, most LLM-based forecasters assume regularly sampled sequences and fixed-step tokenization, which obscure actual time gaps, missing intervals, and asynchronous node updates. ISTS-PLM~\cite{ISTS-PLM} introduces interval-aware time encodings for irregular time series, but remains token-discrete and primarily targets individual sequences. Extending such methods to graph-based forecasting is therefore non-trivial because spatial coupling and temporal irregularity must be handled simultaneously.

Neural ODEs offer a natural alternative by evolving latent states according to actual elapsed time, reconstructing trajectories between sparse observations, and supporting queries at arbitrary timestamps. Their integration with frozen LLMs nevertheless raises two challenges. First, ODE encoders produce long trajectories containing many locally smooth and redundant neighboring states. Directly exposing all states increases the token budget and attention cost while potentially diluting informative transitions. Second, compressed numerical tokens remain misaligned with the representation space learned through language pretraining. Direct concatenation neither explicitly retrieves evidence relevant to a prediction request nor controls how strongly that evidence modifies frozen hidden representations. Effective integration therefore requires compact memory construction and controlled, prediction-aware evidence injection.

To address these challenges, we propose \model{}, an ODE-enhanced frozen-LLM framework for irregular spatio-temporal forecasting. A graph-aware ODE encoder reconstructs asynchronous and partially observed graph sequences as a continuous-time latent trajectory without imposing a regular temporal grid. A Fixed-Budget Perceiver Resampler compresses this trajectory into dynamic memory tokens, using a dynamics-aware attention bias and explicit time labels to preserve informative changes and their temporal locations. In parallel, compact statistical descriptors are encoded and resampled into context memory tokens that provide complementary global information. A dual-source gated cross-attention module then injects both memories into selected layers of the frozen LLM. For each requested prediction timestamp, the LLM prompt contains a dedicated query-slot token, whose final hidden state is mapped to the corresponding forecast. Experiments against 11 baselines on three urban datasets and two physical-dynamics benchmarks show competitive overall performance, with more pronounced advantages under sparse or dynamically complex observations. On NYCbike zero-shot, \model{} reduces token usage by 89.2\% and achieves a $7.7\times$ inference speedup over UrbanGPT while maintaining strong generalization to unseen regions.

Our main contributions are as follows.

\begin{itemize}
    \item We formulate irregular spatio-temporal graph forecasting with a frozen LLM and identify temporal irregularity, representation-space alignment, and context-length constraints as the central challenges.

    \item We propose \textbf{\model}, which combines a graph-aware ODE encoder with a Fixed-Budget Perceiver Resampler to reconstruct continuous-time graph dynamics and compress them into dynamic memory tokens under a bounded token budget.

    \item We introduce dual-source gated cross-attention with gated feed-forward refinement to integrate dynamic and context memories into a frozen LLM in a controlled manner.

    \item Extensive experiments on urban and physical-dynamics benchmarks demonstrate competitive forecasting performance, improved robustness under sparse irregular observations, and strong zero-shot generalization to unseen regions.
\end{itemize}

\section{Problem Formulation}

\begin{figure*}[t]
    \centering
    \captionsetup{skip=3pt}
    \includegraphics[width=1.0\textwidth]{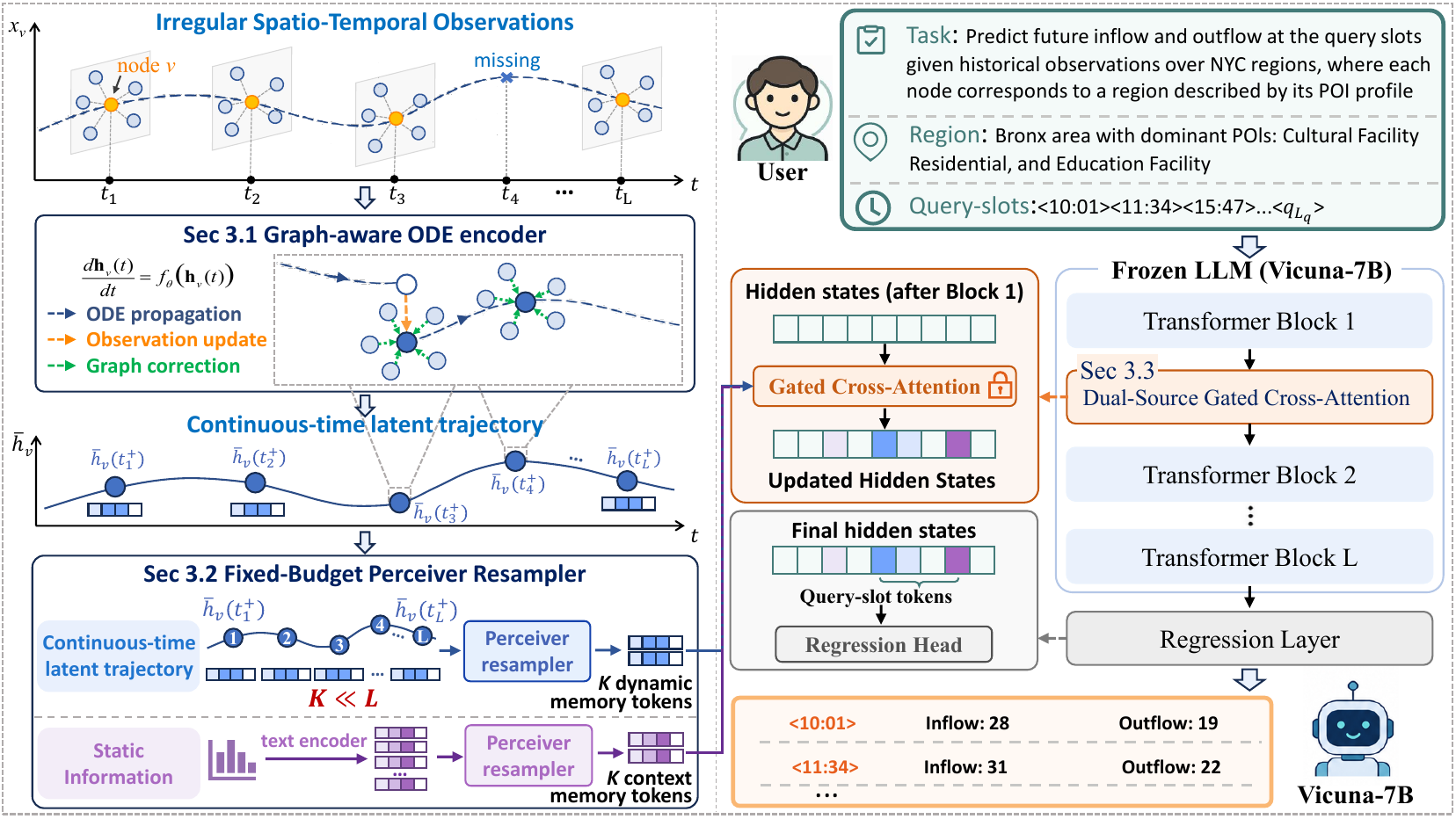}
    \caption{\textbf{Framework of LLMODE.} LLMODE constructs two memory streams to condition a frozen LLM backbone. The graph-aware ODE encoder first reconstructs a continuous-time latent trajectory, which is resampled into a fixed number of dynamic memory tokens. In parallel, compact statistical descriptors are encoded and resampled into context memory tokens. Prompt hidden states retrieve information from both memories through
    dual-source gated cross-attention, and the final query-slot states are
    mapped to forecasts.}
    \label{fig:framework_overview}
\end{figure*}

Let $\mathcal{G}=(\mathcal{V},\mathcal{E})$ denote a fixed graph, where
$\mathcal{V}$ and $\mathcal{E}$ represent the node and edge sets,
respectively, and $|\mathcal{V}|=N$ is the number of nodes. Each node
$v\in\mathcal{V}$ is associated with a $D$-dimensional observation vector
$\mathbf{x}_v(t)\in\mathbb{R}^{D}$ at continuous time $t$, where $D$ denotes
the number of observed features.

Instead of observing all nodes on a shared regular time grid, we collect an
irregular observation set:
\begin{equation}
\mathcal{D}_{\mathrm{obs}}
=
\{
(v,t_{v,i},\mathbf{x}_v(t_{v,i}))
\},
\end{equation}
where $t_{v,i}$ denotes the $i$-th observation timestamp of node $v$ and
$\mathbf{x}_v(t_{v,i})$ is the corresponding observed feature vector. The
observation intervals $t_{v,i+1}-t_{v,i}$ are non-uniform, and different nodes
may have different timestamp sets.

Given a set of query timestamps
$\mathcal{Q}_v=\{t_{v,n}^{(q)}\}_{n=1}^{L_q}$ for node $v$,
where $L_q$ denotes the number of query points, the goal is
to jointly predict a multivariate target vector at each arbitrary
query time:
\begin{equation}
\widehat{\mathbf{y}}_v\left(t_{v,n}^{(q)}\right)
=
f\left(\mathcal{G},\mathcal{D}_{\mathrm{obs}},v,
t_{v,n}^{(q)}\right)
\in \mathbb{R}^{D},
\end{equation}
where $D$ denotes the number of jointly predicted target
variables, and
$\widehat{\mathbf{y}}_v(t_{v,n}^{(q)})$
is the predicted target vector of node $v$ at query time
$t_{v,n}^{(q)}$.
\section{Method}


As illustrated in Figure~\ref{fig:framework_overview}, LLMODE comprises a graph-aware ODE encoder that reconstructs continuous-time graph dynamics, a Fixed-Budget Perceiver Resampler that compresses dynamic trajectories and context descriptors into fixed-size memories, and a dual-source gated cross-attention module that injects both memories into the frozen LLM for forecasting.

\subsection{Graph-aware ODE Encoder}
\label{sec:latent_dynamics}

Irregular sampling and missing observations create nonuniform time intervals
during which the underlying system continues to evolve without direct
measurements. We therefore represent each node $v$ by a continuous latent
state governed by
\begin{equation}
\frac{d\mathbf{h}_v(t)}{dt}
=
f_{\theta}\!\left(\mathbf{h}_v(t)\right),
\label{eq:latent_ode}
\end{equation}
where $\mathbf{h}_v(t)\in\mathbb{R}^{d_{\mathrm{ode}}}$ denotes the latent state of node
$v$, and $f_{\theta}$ is a learnable vector field instantiated using a
GRU-ODE parameterization. An ODE solver propagates the latent state over the
actual elapsed interval between observations.

When an observation $\mathbf{x}_{v,i}\in\mathbb{R}^{D}$ arrives at time
$t_i$, the ODE solver first produces the propagated state
$\mathbf{h}_v(t_i^-)$. Because continuous propagation may gradually deviate
from the observed trajectory, we then use the newly arrived observation to
correct this state:
\begin{equation}
\mathbf{h}_v(t_i^{+})
=
\mathrm{GRU}\!\left(
\mathbf{h}_v(t_i^{-}),
\Phi\!\left(\mathbf{x}_{v,i},\mathbf{h}_v(t_i^{-})\right)
\right),
\label{eq:observation_jump}
\end{equation}
where the superscripts $-$ and $+$ denote the latent states immediately
before and after observation correction, respectively. The observation
encoder $\Phi$ constructs the correction signal from the new observation and
its discrepancy from the current latent prediction.

To capture structural interactions, the corrected node states
$\{\mathbf{h}_v(t_i^+)\}_{v\in\mathcal{V}}$ are further refined through a
gated graph-based residual update. The resulting graph-aware state
$\bar{\mathbf{h}}_v(t_i^+)$ is used as the initial state for continuous
propagation over the next interval. The detailed GRU-ODE parameterization,
observation encoding, and graph-aware correction are provided in
Appendix~\ref{app:GRU-ODE parameterization}--\ref{app:graph-aware correction}.

At a set of solver timestamps $\{\tau_l\}_{l=1}^{L}$, the states of all nodes
are collected as $
\bar{\mathbf{H}}(\tau_l)
\in\mathbb{R}^{N\times d_{\mathrm{ode}}}$ and stacked along the
temporal dimension to construct the continuous-time latent trajectory
\begin{equation}
\mathbf{H}_{\mathrm{dyn}}
=
\left[
\bar{\mathbf{H}}(\tau_1);
\ldots;
\bar{\mathbf{H}}(\tau_L)
\right]
\in\mathbb{R}^{N\times L\times d_{\mathrm{ode}}}.
\label{eq:continuous_latent_trajectory}
\end{equation}
The resulting trajectory captures the continuous temporal evolution and
structural interactions of all nodes and serves as the input to the
subsequent fixed-budget resampler.

\subsection{Fixed-Budget Perceiver Resampler}
\label{sec:token_compression}

Given an irregularly sampled input trajectory, the graph-aware ODE encoder
uses its smallest observation gap as the ODE integration step and records the
propagated latent states as $\mathbf{H}_{\mathrm{dyn}}
\in\mathbb{R}^{N\times L\times d_{\mathrm{ode}}}$. 
These states are projected into the resampler space
$\mathbb{R}^{d_{\mathrm{mem}}}$ before resampling. Although the reconstructed
trajectory provides fine-grained dynamic states, directly injecting all
$L$ states is computationally expensive and may dilute informative changes
because adjacent states can be highly redundant.

We therefore use a Perceiver Resampler to distill each reconstructed latent
trajectory into a fixed number of dynamic memory tokens. For clarity, we omit the node index $v$ in the following derivation and describe the resampling for a single node. Specifically, the
resampler uses $K$ learnable latent queries to attend to the input sequence
and outputs $\{\mathbf{z}_k\}_{k=1}^{K}$, where
$\mathbf{z}_k\in\mathbb{R}^{d_{\mathrm{mem}}}$.
Rapidly changing intervals generally contain more informative dynamic
evidence. However, standard attention alone does not explicitly encourage
the resampler to preserve states from these intervals. We therefore introduce
the magnitude of the latent dynamics directly into the attention logits:
\begin{equation}
\alpha_{k,i}
=
\operatorname{softmax}_{i}\!\left(
\frac{\mathbf{q}_{k}^{\top}\mathbf{k}_{i}}{\sqrt{d_{\mathrm{mem}}}}
+
\beta\,
\operatorname{Norm}\!\left(
\left\|\dot{\mathbf{h}}(t_i)\right\|_2
\right)
\right).
\label{eq:dynamics_attention}
\end{equation}
Here, $\alpha_{k,i}$ is the attention weight from the $k$-th latent query to
the state at timestamp $t_i$;
$\mathbf{q}_k\in\mathbb{R}^{d_{\mathrm{mem}}}$ is a
learnable latent query, and $\mathbf{k}_i\in\mathbb{R}^{d_{\mathrm{mem}}}$ is the key
representation obtained from the latent state $\mathbf{h}(t_i)$ through a
learnable projection; $d_{\mathrm{mem}}$ denotes the resampler and memory-token
dimension;
$\dot{\mathbf{h}}(t_i)=f_{\theta}(\mathbf{h}(t_i))$ represents the
instantaneous latent change given by the ODE vector field;
$\operatorname{Norm}(\cdot)$ denotes normalization over the latent trajectory;
and $\beta$ controls the strength of the dynamics-aware bias.
This bias encourages the resampler to retain states from intervals with more
pronounced changes while preserving content-based attention.

The downstream forecasting prompt specifies the timestamps to be predicted.
To align the distilled dynamic memory content with these query times, we
assign each resampled vector an explicit time label derived from the
same attention weights. The time label is fused with $\mathbf{z}_k$ to form
the time-aware representation $\tilde{\mathbf{z}}_k$.
Collecting the $K$ tokens yields the dynamic memory
\begin{equation}
\tilde{\mathbf{Z}}
=
[\tilde{\mathbf{z}}_1;\ldots;\tilde{\mathbf{z}}_K]
\in\mathbb{R}^{K\times d_{\mathrm{mem}}}.
\end{equation}
The detailed resampling and time-labeling process is provided in
Appendix~\ref{app:detailed resampling}--~\ref{app:time-labeling}.



\begin{table*}[!t]
\centering
\caption{Main results on NYC and physics benchmarks, measured by MAE and MSE. Results are averaged over $5$ independent runs. Overall, \model{} achieves an average relative improvement of 6.3\% across all datasets and evaluation metrics. $*$ indicates the statistically significant improvements over the best baseline, with $p$-value smaller than $0.001$.}
\label{tab:main_results}
\vspace{-4pt}
\resizebox{\textwidth}{!}{
\begin{tabular}{ccccccccccccccccccccccc}
\toprule
\multirow{4}{*}{Method} 
& \multicolumn{4}{c}{NYCtaxi} 
& \multicolumn{4}{c}{NYCbike} 
& \multicolumn{4}{c}{NYCcrime}
& \multicolumn{4}{c}{Springs}
& \multicolumn{4}{c}{Charged}\\
\cmidrule(lr){2-5}\cmidrule(lr){6-9}\cmidrule(lr){10-13}\cmidrule(lr){14-17}\cmidrule(lr){18-21}
& \multicolumn{2}{c}{inflow} 
& \multicolumn{2}{c}{outflow} 
& \multicolumn{2}{c}{inflow} 
& \multicolumn{2}{c}{outflow} 
& \multicolumn{2}{c}{robbery}
& \multicolumn{2}{c}{burglary} 
& \multicolumn{2}{c}{X}
& \multicolumn{2}{c}{Y} 
& \multicolumn{2}{c}{X}
& \multicolumn{2}{c}{Y} \\
\cmidrule(lr){2-3}\cmidrule(lr){4-5}\cmidrule(lr){6-7}\cmidrule(lr){8-9}\cmidrule(lr){10-11}\cmidrule(lr){12-13}\cmidrule(lr){14-15}\cmidrule(lr){16-17}\cmidrule(lr){18-19}\cmidrule(lr){20-21}
& MAE & MSE & MAE & MSE & MAE & MSE & MAE & MSE & MAE & MSE & MAE & MSE & MAE & MSE & MAE & MSE & MAE & MSE & MAE & MSE \\


\midrule
TGCN  
& 0.1145 & 0.0367 & 0.0875 & 0.0331 
& 0.2211 & 0.1572 & 0.1506 & 0.0657 
& 0.9970 & 1.0849 & 0.6746 & 1.0029 
& 0.1971 & 0.0741 & 0.1934 & 0.0708 
& 0.4706 & 0.3038 & 0.4710 & 0.3042 \\
STSGCN    
& 0.0847 & 0.0286 & 0.0704 & 0.0349 
& 0.1951 & 0.1734 & 0.1669 & 0.1119 
& 0.9162 & 1.5371 & 0.6293 & 0.8300 
& 0.3306 & 0.1686 & 0.3114 & 0.1754 
& 0.3804 & 0.2153 & 0.4152 & 0.2879 \\
MTGNN      
& 0.0371 & 0.0143 & 0.0337 & 0.0365 
& \secondbest{0.0786} & \secondbest{0.0263} & \secondbest{0.0824} & \secondbest{0.0265} 
& 0.8252 & 1.1408 & 0.6283 & 0.7857 
& \secondbest{0.1053} & \secondbest{0.0237} & \secondbest{0.1086} & \secondbest{0.0249} 
& \secondbest{0.1410} & 0.0533 & 0.1482 & 0.0535 \\
BiTGraph  
& 0.0731 & 0.0305 & 0.0540 & 0.0301 
& 0.0837 & 0.0298 & 0.0830 & 0.0275 
& 0.8555 & 1.2376 & 0.5321 & 0.4988 
& 0.1271 & 0.0312 & 0.1290 & 0.0321 
& 0.1501 & 0.0578 & 0.1527 & 0.0593 \\
FourierGNN    
& 0.1234 & 0.0246 & 0.1026 & 0.0187 
& 0.1533 & 0.0734 & 0.1523 & 0.0626 
& 0.4074 & 0.8993 & 0.2666 & 0.2899 
& 0.2108 & 0.0767 & 0.2127 & 0.0788 
& 0.2305 & 0.1000 & 0.2309 & 0.1003 \\

\midrule
ISTS-PLM 
& \textbf{0.0329} & \secondbest{0.0068} & 0.0263 & \secondbest{0.0059}  
& 0.1061 & 0.0461 & 0.1077 & 0.0488 
& 0.4271 & 0.4055 & 0.3040 & 0.2143 
& 0.1513 & 0.0427 & 0.1505 & 0.0423 
& 0.1524 & 0.0569 & 0.1539 & 0.0576 \\
UrbanGPT        
& 0.0359 & 0.0080 & \secondbest{0.0262} & 0.0068 
& 0.0787 & 0.0301 & 0.0833 & 0.0292 
& 0.9336 & 1.7088 & 0.6192 & 0.8341 
& 0.1519 & 0.0420 & 0.1514 & 0.0424 
& 0.1633 & 0.0702 & 0.1746 & 0.0908 \\
GPT4TS               
& 0.0424 & 0.0260 & 0.0312 & 0.0209 
& 0.1592 & 0.2072 & 0.1636 & 0.2129 
& 0.3350 & 0.2578 & \secondbest{0.2398} & 0.1229 
& 0.4549 & 0.2826 & 0.4549 & 0.2838 
& 0.4996 & 0.3447 & 0.5009 & 0.3469 \\

\midrule
GRU-ODE  
& 0.0956 & 0.0645 & 0.0708 & 0.0607 
& 0.3100 & 0.4081 & 0.3443 & 0.4964 
& 0.3678 & 0.4161 & 0.2483 & 0.1815 
& 0.3132 & 0.1664 & 0.3080 & 0.1615 
& 0.3001 & 0.1543 & 0.2888 & 0.1399 \\
tPatchGNN     
& 0.0338 & 0.0082 & 0.0347 & 0.0095 
& 0.0800 & 0.0476 & 0.0849 & 0.0484 
& \secondbest{0.3052} & \secondbest{0.1428} & 0.2483 & \secondbest{0.1119}
& 0.1391 & 0.0370 & 0.1384 & 0.0367 
& 0.1439 & \secondbest{0.0504} & \secondbest{0.1497} & \secondbest{0.0504} \\
ViTST     
& 0.3548 & 0.1718 & 0.2938 & 0.1377 
& 0.4371 & 0.6891 & 0.4542 & 0.7136 
& 0.4582 & 0.5084 & 0.3267 & 0.2325 
& 0.4519 & 0.2751 & 0.4524 & 0.2767 
& 0.4842 & 0.3174 & 0.4849 & 0.3181 \\

\midrule
\textbf{\model} 
& \secondbest{0.0336} & \textbf{0.0066} & \textbf{0.0252} & \textbf{0.0057} 
& \textbf{0.0767}$^*$ & \textbf{0.0243}$^*$ & \textbf{0.0817} & \textbf{0.0242}$^*$ 
& \textbf{0.3050} & \textbf{0.1396}$^*$ & \textbf{0.2394} & \textbf{0.0790$^*$} 
& \textbf{0.0973}$^*$ & \textbf{0.0191}$^*$ & \textbf{0.0986}$^*$ & \textbf{0.0196}$^*$ 
& \textbf{0.1402} & \textbf{0.0483}$^*$ & \textbf{0.1461}$^*$ & \textbf{0.0493}$^*$ \\
\bottomrule
\end{tabular}}
\end{table*}

\subsection{Dual-Source Gated Cross-Attention}
\label{sec:gca}

In addition to the dynamic memory $\widetilde{\mathbf Z}$ extracted from the continuous-time latent trajectory, we construct a context memory $\mathbf P$ from the observed history. The observations are summarized as compact statistical descriptions, encoded by a language encoder, and compressed by a resampler into $K$ context memory tokens, $\mathbf{P}\in\mathbb{R}^{K\times d_{\mathrm{mem}}}$. While $\widetilde{\mathbf Z}$ captures fine-grained temporal dynamics, $\mathbf P$ provides complementary global statistics. Details of the context-memory construction are provided in Appendix~\ref{app:context_memory}.

The two memories are injected into the frozen LLM through dual-source gated cross-attention, avoiding the uncontrolled mixing caused by direct prompt concatenation. A forecasting prompt specifies the task and arbitrary query timestamps. Representative prompts are provided in Appendix~\ref{app:prompt_details}. After a Transformer layer, its hidden states $\mathbf Y$ are updated as follows:
\begin{equation}
\boldsymbol{\Delta}_{\mathrm{mem}}
=
\lambda \operatorname{CrossAttn}
\left(
\mathbf{Y},
\widetilde{\mathbf{Z}}
\right)
+
(1-\lambda)
\operatorname{CrossAttn}
\left(
\mathbf{Y},
\mathbf{P}
\right).
\label{eq:memory_fusion}
\end{equation}

\begin{equation}
\widehat{\mathbf{Y}}
=
\mathbf{Y}
+
\alpha_{\mathrm{attn}}
\boldsymbol{\Delta}_{\mathrm{mem}}.
\label{eq:gated_cross_attention}
\end{equation}

\begin{equation}
\mathbf{Y}^{\prime}
=
\widehat{\mathbf{Y}}
+
\alpha_{\mathrm{ffn}}
\operatorname{FFN}
\left(
\widehat{\mathbf{Y}}
\right),
\label{eq:gated_ffn}
\end{equation}
where
\begin{equation}
\begin{aligned}
\alpha_{\mathrm{attn}}
&=
\tanh\left(g_{\mathrm{attn}}\right),\\
\alpha_{\mathrm{ffn}}
&=
\tanh\left(g_{\mathrm{ffn}}\right).
\end{aligned}
\label{eq:gate_gains}
\end{equation}
Here, $\mathbf{Y},\widehat{\mathbf{Y}},\mathbf{Y}^{\prime}\in\mathbb{R}^{L_p\times d_{\mathrm{LLM}}}$ denote the prompt-token hidden states before injection, after cross-attention, and after feed-forward refinement, respectively. Among the $L_p$ prompt tokens, $L_q$ serve as query slots. $\operatorname{CrossAttn}(\mathbf Y,\mathbf M)$ projects the prompt states
and memory tokens into a shared attention space, using them as queries and
keys/values, respectively, and returns an update in
$\mathbb{R}^{L_p\times d_{\mathrm{LLM}}}$, while $\lambda$ balances the two memories. The learnable parameters $g_{\mathrm{attn}}$ and $g_{\mathrm{ffn}}$ are initialized to zero, making the module an identity mapping at the beginning of training.

For Vicuna-7B, we insert two such modules after Transformer Layers 1 and 16. 
The updated states are propagated through the remaining layers. Each query-slot token corresponds to a prediction timestamp, and its final-layer hidden state is extracted and projected by a lightweight regression head to produce the corresponding forecast. The model is trained using MSE, with the complete training configuration provided in Appendix~\ref{app:training_configuration}.
\providecommand{\tbd}{\textbf{TBD}}

\section{Experiments}

Our experiments are designed to answer the following questions:

\textbf{Q1: Overall effectiveness.}
Can LLMODE achieve competitive forecasting performance across both real-world urban datasets and physical-dynamics benchmarks?

\textbf{Q2: Unseen-region generalization.}
Can LLMODE transfer to spatial regions that are entirely unseen during training and forecast their future states without additional adaptation?

\textbf{Q3: Necessity of continuous-time modeling.}
Does continuous-time encoding become increasingly important as observations become sparser, compared with discrete spatio-temporal encoders and existing baselines?

\textbf{Q4: Effectiveness of fixed-budget dynamic evidence.}
Can a small number of adaptively selected dynamic states preserve the forecasting-relevant information in the continuous trajectory more effectively than exposing the full trajectory or applying simple compression?

\textbf{Q5: Contribution and utilization of the frozen LLM.}
First, do pretrained frozen representations provide benefits beyond a prediction network or a Transformer trained from scratch? Second, does the frozen backbone effectively use the injected dynamic and contextual evidence, and does controlled gated injection improve this utilization?


We first describe the experimental setup and then address Q1--Q5, followed by a qualitative case study. 

\subsection{Experimental Setup}

\textbf{Datasets and protocols.} We use NYCtaxi, NYCbike, and NYCcrime for urban forecasting, and Springs and Charged for particle dynamics. NYCtaxi/NYCbike use 24-hour histories to predict 12 hours; NYCcrime uses 96 days to predict 48 days. The supervised setting evaluates training regions, while the zero-shot setting holds out 80 NYC regions. Springs and Charged predict future 2D particle positions from asynchronous partial observations. Dataset construction and preprocessing details are provided in Appendix ~\ref{app:datasets_construction}.

\textbf{Baselines.} We compare against classical spatio-temporal models, including TGCN, STSGCN, MTGNN, BiTGraph, and FourierGNN; LLM-based forecasters, including ISTS-PLM, UrbanGPT, and GPT4TS; and irregular-aware methods, including GRU-ODE, tPatchGNN, and ViTST~\cite{TGCN,STSGCN,MTGNN,BiTGraph,FourierGNN,ISTS-PLM,UrbanGPT,GPT4TS,GRU-ODE,tPatchGNN,ViTST}. Regular-grid baselines use linearly interpolated inputs. Detailed descriptions of the baselines are provided in Appendix ~\ref{app:baselines}.

\textbf{Implementation.} We use frozen Vicuna-7B~\cite{vicuna7b} as the default backbone. Detailed model configurations, optimization settings, and hardware information are provided in Appendix~\ref{app:implementation}.
\subsection{Q1: Overall Forecasting Effectiveness}

To answer Q1, Table~\ref{tab:main_results} compares LLMODE with conventional spatio-temporal models, irregular-aware methods, and LLM-based forecasters on three real-world urban datasets and two physical-dynamics benchmarks.

Overall, LLMODE achieves competitive or superior performance across all five benchmarks. The strongest gains are observed on NYCcrime, Springs, and Charged, where observations are more event-driven, sparse, or governed by complex interactions. This result answers Q1 affirmatively and indicates that the complete framework is not restricted to either urban data or physical systems. The smaller gains on the denser mobility datasets are further analyzed
in the Appendix ~\ref{app:distribution_shift_analysis}.


\subsection{Q2: Unseen-Region Generalization}

\begin{table}[!t]
\centering
\caption{Zero-shot performance on NYC benchmarks, measured by MAE and MSE. Results are averaged over inflow and outflow for NYCtaxi and NYCbike, and over robbery and burglary for NYCcrime. Average results on $5$ independent runs are reported. Numbers below our results indicate relative changes compared with the
strongest baseline for each metric. Averaged over all
six metrics, \model{} achieves an average relative improvement of 11.4\%.
}
\label{tab:zeroshot_nyc}
\vspace{-8pt}
\resizebox{1.0\columnwidth}{!}{
\begin{tabular}{lcccccc}
\toprule
\multirow{2}{*}{Method}
& \multicolumn{2}{c}{NYCtaxi} 
& \multicolumn{2}{c}{NYCbike} 
& \multicolumn{2}{c}{NYCcrime} \\

\cmidrule(lr){2-3}\cmidrule(lr){4-5}\cmidrule(lr){6-7}
& MAE & MSE & MAE & MSE & MAE & MSE \\

\midrule
TGCN  
& 0.3233 & 0.6351 
& 0.2457 & 0.2105 
& 0.7266 & 2.6862 \\
STSGCN    
& 0.2758 & 0.4385 
& 0.2221 & 0.2723 
& 0.7431 & 2.8640 \\
MTGNN      
& 0.0878 & 0.0748 
& 0.0769 & 0.0357 
& 0.5574 & 1.7864 \\
BiTGraph  
& 0.0964 & \secondbest{0.0563} 
& 0.0880 & 0.0415 
& 0.4703 & 1.5098 \\
FourierGNN    
& 0.1538 & 0.0628 
& 0.1528 & 0.0680 
& 0.5304 & 1.4962 \\

\midrule
ISTS-PLM 
& 0.1042 & 0.0931 
& 0.0822 & 0.0371 
& 0.2041 & \secondbest{0.3421} \\
UrbanGPT        
& \textbf{0.0798} & 0.0634 
& 0.0778 & \secondbest{0.0292} 
& 0.7418 & 2.0739 \\
GPT4TS                
& 0.1210 & 0.1524 
& 0.1017 & 0.0834 
& 0.2885 & 0.8108 \\

\midrule
GRU-ODE  
& 0.2510 & 0.4434 
& 0.2348 & 0.2410 
& 0.4219 & 1.6564 \\
tPatchGNN     
& 0.0830 & 0.0577 
& \secondbest{0.0765} & 0.0323 
& \secondbest{0.1644} & 0.3863 \\
ViTST     
& 0.4427 & 0.5576 
& 0.3676 & 0.3013 
& 0.8560 & 2.7815 \\

\midrule
\textbf{LLMODE (Ours)}
& \shortstack{\underline{0.0801}\\
  {\scriptsize\color{degrade}{$\uparrow 0.4\%$}}}
& \shortstack{\textbf{0.0396}\\
  {\scriptsize\color{improve}{$\downarrow 29.7\%$}}}
& \shortstack{\textbf{0.0745}\\
  {\scriptsize\color{improve}{$\downarrow 2.6\%$}}}
& \shortstack{\textbf{0.0260}\\
  {\scriptsize\color{improve}{$\downarrow 11.0\%$}}}
& \shortstack{\textbf{0.1527}\\
  {\scriptsize\color{improve}{$\downarrow 7.1\%$}}}
& \shortstack{\textbf{0.2791}\\
  {\scriptsize\color{improve}{$\downarrow 18.4\%$}}}
\\
\bottomrule
\end{tabular}}
\end{table}
To answer Q2, we evaluate LLMODE under the zero-shot protocol on the NYC benchmarks. The model is trained on 80 regions and directly tested on another 80 unseen regions without fine-tuning. This setting tests whether LLMODE can infer future dynamics for unseen regions from their observed histories, rather than relying on region-specific patterns
learned during training.

As shown in Table~\ref{tab:zeroshot_nyc}, LLMODE achieves the best performance on five of the six metrics. The gains are particularly clear on the much sparser NYCcrime dataset, indicating effective generalization to unseen regions under sparse and irregular observations. The contribution of the frozen LLM to this transferability is further analyzed in Q5.

\subsection{Q3: Necessity of Continuous-Time Modeling}

\begin{figure}[htbp]
\centering
\includegraphics[width=.5\linewidth]{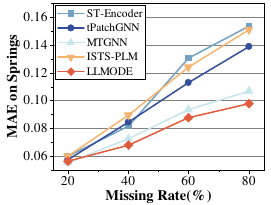}\hfill
\includegraphics[width=.5\linewidth]{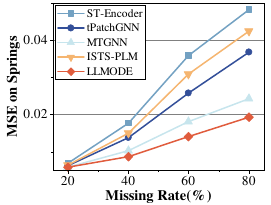}
\caption{Forecasting performance under different missing rates on the Springs system.}
\label{fig:missing_ratio}
\end{figure}

To answer Q3, Figure~\ref{fig:missing_ratio} compares LLMODE with representative methods under controlled missing rates. ST-Encoder replaces the graph-aware ODE encoder with the discrete spatio-temporal encoder used in UrbanGPT, while keeping all other components unchanged. The remaining methods include representative regular- and irregular-sampling baselines.

As the missing rate increases, ST-Encoder degrades more rapidly than LLMODE, indicating the limitation of discrete encoders under sparse observations. MTGNN remains relatively robust due to its adaptive graph and multi-scale temporal modeling, whereas tPatchGNN and ISTS-PLM also show increasing errors. In contrast, LLMODE consistently achieves the lowest errors, with a widening advantage at higher missing rates. These results support the importance of continuous-time modeling for reconstructing dynamics over long unobserved intervals. Solver and step-size sensitivity analyses are provided in Appendix ~\ref{app:numerical_sensitivity}.

\subsection{Q4: Effectiveness of Fixed-Budget Dynamic Evidence}

\begin{figure}[htbp]
\centering
\includegraphics[width=0.85\linewidth]{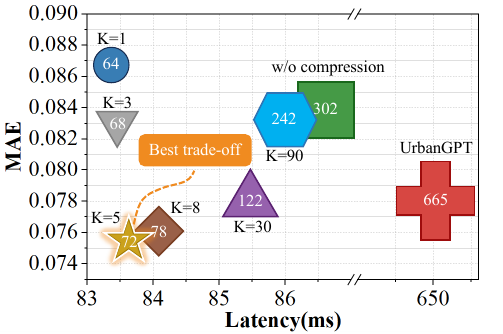}
\caption{Token-budget efficiency analysis on NYCbike zero-shot. Numbers inside the markers indicate token counts.}
\label{fig:token_budget}
\end{figure}

To answer Q4, we examine both the token budget and the trajectory-state selection strategy. The continuous-time encoder reconstructs each trajectory into $L=120$ latent states, and Figure~\ref{fig:token_budget} compares different fixed-budget settings and full-trajectory exposure, with UrbanGPT included as an LLM-based reference. 
Compared with UrbanGPT, LLMODE reduces the reported token budget by nearly 90\% and inference latency by approximately 87\%, while achieving better zero-shot accuracy. An extremely small budget ($K=1$) loses substantial trajectory information, whereas $K=5$ and $K=8$ achieve the strongest accuracy--efficiency trade-off. Across datasets and
task settings, these two budgets consistently perform best, indicating that a small fixed token budget is robust
across different forecasting settings. Larger budgets and full-trajectory exposure yield no consistent gains, suggesting that forecasting-relevant information is concentrated in a small subset of states. Detailed results, together with the inference latency of representative non-LLM methods for reference, are provided in Appendix~\ref{app:token_budget}. Figure~\ref{fig:evidence_selection} further compares learned resampling with random selection and linear projection. Learned resampling consistently performs better, indicating that adaptive state selection is more effective than simple sequence reduction. Removing either the dynamics-aware bias or explicit time labels also degrades performance, confirming that compact dynamic memory must preserve both salient state changes and their temporal locations.



\begin{figure}[htbp]
\centering
\includegraphics[width=.5\linewidth]{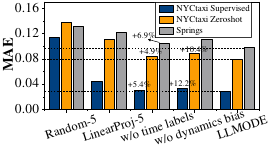}\hfill
\includegraphics[width=.5\linewidth]{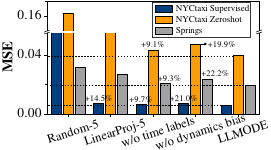}
\caption{Comparison of different evidence-selection mechanisms under a fixed token budget. The left and right panels report MAE and MSE, respectively.}
\label{fig:evidence_selection}
\end{figure}

\subsection{Q5: Frozen-LLM Contribution and Evidence Utilization}

To answer Q5, we compare the frozen LLM with an MLP and a randomly initialized six-layer Transformer to examine the benefits of pretrained representations beyond trainable models. We further evaluate different frozen backbones, evidence injection mechanisms, and evidence contents to assess the
robustness of the backbone and the effectiveness of evidence utilization.

\begin{table}[t]
\centering
\scriptsize
\caption{Contribution and robustness of the frozen LLM, evaluated by MAE and MSE. Percentage changes next to each result are computed relative to the default Vicuna-7B backbone.}
\label{tab:llm_backbone}
\setlength{\tabcolsep}{3pt}

\resizebox{\columnwidth}{!}{%
\begin{tabular}{@{}lcccc@{}}
\toprule
\multirow{2}{*}{Model}
& \multicolumn{2}{c}{NYCtaxi Supervised}
& \multicolumn{2}{c}{NYCtaxi Zero-shot} \\
\cmidrule(lr){2-3}
\cmidrule(lr){4-5}
& MAE & MSE & MAE & MSE \\

\midrule
\multicolumn{5}{l}{\textit{Pretraining and capacity controls}} \\

Ours w/ MLP
& 0.0317{\tiny\color{degrade}{$\uparrow7.2\%$}}
& 0.0071{\tiny\color{degrade}{$\uparrow12.6\%$}}
& 0.1131{\tiny\color{degrade}{$\uparrow41.2\%$}}
& 0.1048{\tiny\color{degrade}{$\uparrow164.6\%$}} \\

Ours w/ random Transformer
& 0.0301{\tiny\color{degrade}{$\uparrow2.4\%$}}
& 0.0065{\tiny\color{degrade}{$\uparrow4.8\%$}}
& 0.0925{\tiny\color{degrade}{$\uparrow15.5\%$}}
& 0.0521{\tiny\color{degrade}{$\uparrow31.6\%$}} \\

\midrule
\multicolumn{5}{l}{\textit{Frozen-backbone robustness}} \\

Best baseline
& 0.0296{\tiny\color{degrade}{$\uparrow0.7\%$}}
& 0.0064{\tiny\color{degrade}{$\uparrow3.2\%$}}
& 0.0798{\tiny\color{improve}{$\downarrow0.4\%$}}
& 0.0563{\tiny\color{degrade}{$\uparrow42.2\%$}} \\

Ours w/ Llama-3.1-8B
& \textbf{0.0288}{\tiny\color{improve}{$\downarrow2.0\%$}}
& \textbf{0.0060}{\tiny\color{improve}{$\downarrow3.2\%$}}
& 0.0804{\tiny\color{degrade}{$\uparrow0.4\%$}}
& 0.0378{\tiny\color{improve}{$\downarrow4.5\%$}} \\

Ours w/ Llama-2-13B
& 0.0296{\tiny\color{degrade}{$\uparrow0.7\%$}}
& 0.0067{\tiny\color{degrade}{$\uparrow8.1\%$}}
& \textbf{0.0793}{\tiny\color{improve}{$\downarrow1.0\%$}}
& \textbf{0.0371}{\tiny\color{improve}{$\downarrow6.3\%$}} \\

\rowcolor{gray!15}
\textbf{Ours w/ Vicuna-7B (default)}
& 0.0294
& 0.0062
& 0.0801
& 0.0396 \\

\bottomrule
\end{tabular}%
}
\end{table}

\paragraph{Contribution of the Frozen LLM.}



As shown in Table~\ref{tab:llm_backbone}, replacing the frozen LLM with trainable alternatives, including an MLP prediction head and a randomly initialized Transformer, achieves comparable supervised performance but substantially degrades zero-shot performance. This suggests
that trainable models can capture forecasting patterns from observed regions, whereas the pretrained LLM provides additional modeling ability that is particularly important for transfer to unseen regions.

Moreover, the three frozen LLM backbones achieve comparable performance on both seen and unseen regions. This consistency shows that the benefit of the frozen LLM is not specific to Vicuna-7B, and that LLMODE can effectively work with different pretrained backbones.



\begin{figure}[htbp]
\centering
\includegraphics[width=0.5\linewidth]{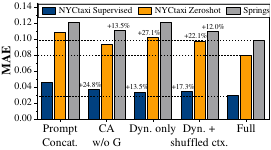}\hfill
\includegraphics[width=0.5\linewidth]{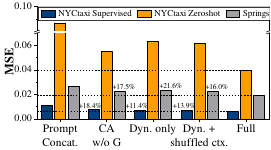}
\caption{Ablation study of evidence access mechanisms and evidence sources.
``Prompt Concat.'' denotes prompt concatenation, ``CA w/o G'' denotes
cross-attention without gating, ``Dyn. only'' uses only dynamic memory,
and ``Dyn. + shuffled ctx.'' uses dynamic memory with randomly shuffled
context memory.}
\label{fig:evidence_injection}
\end{figure}

\begin{figure*}[t]
\centering
\includegraphics[width=.24\textwidth]{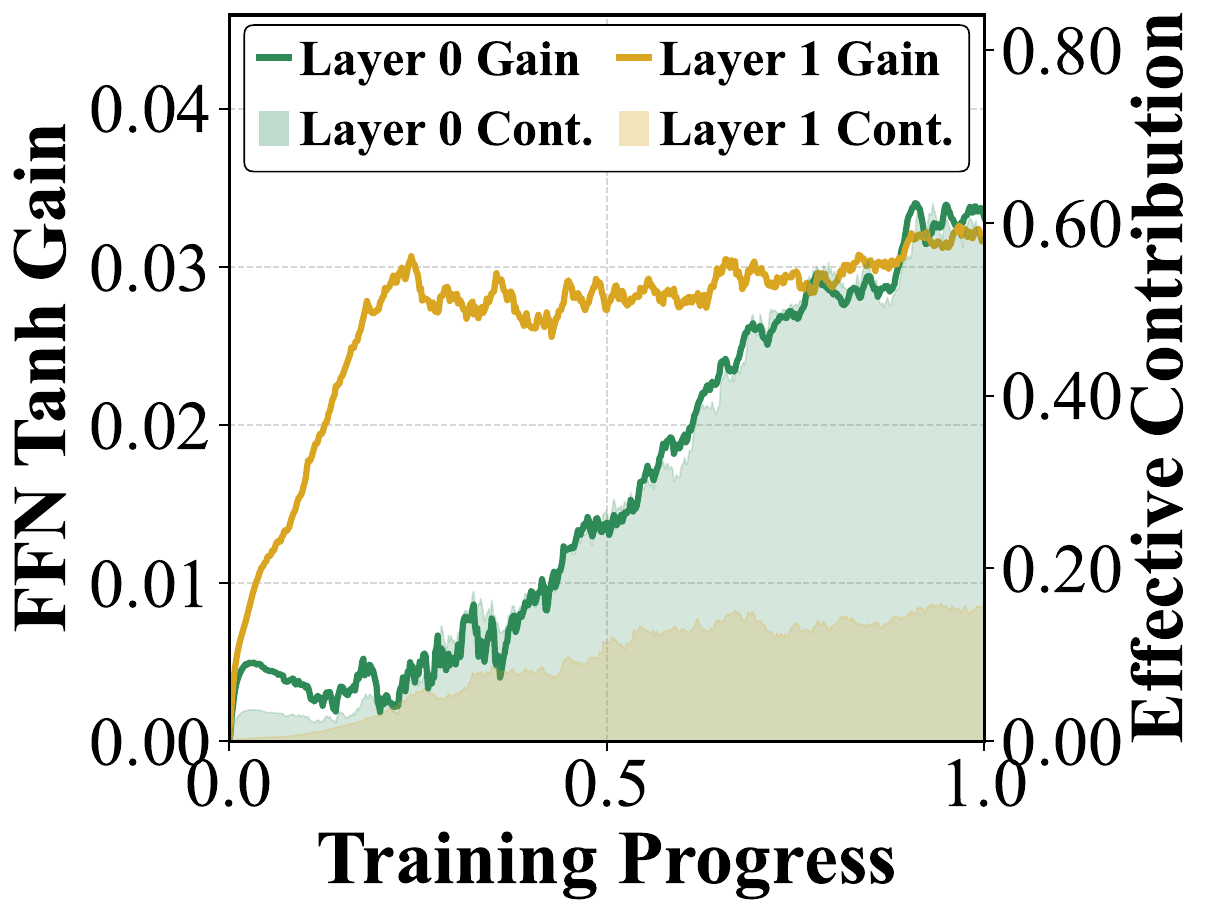}\hfill
\includegraphics[width=.24\textwidth]{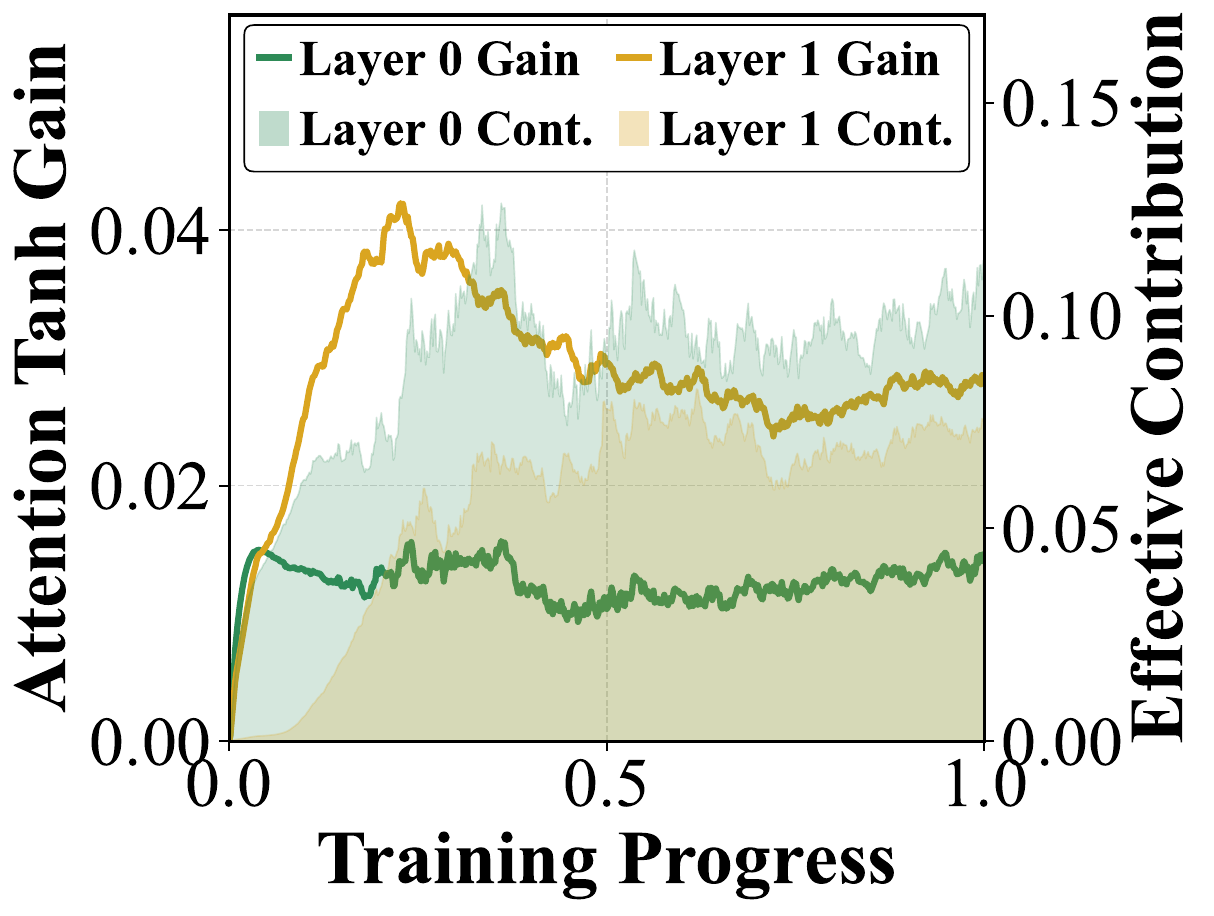}\hfill
\includegraphics[width=.24\textwidth]{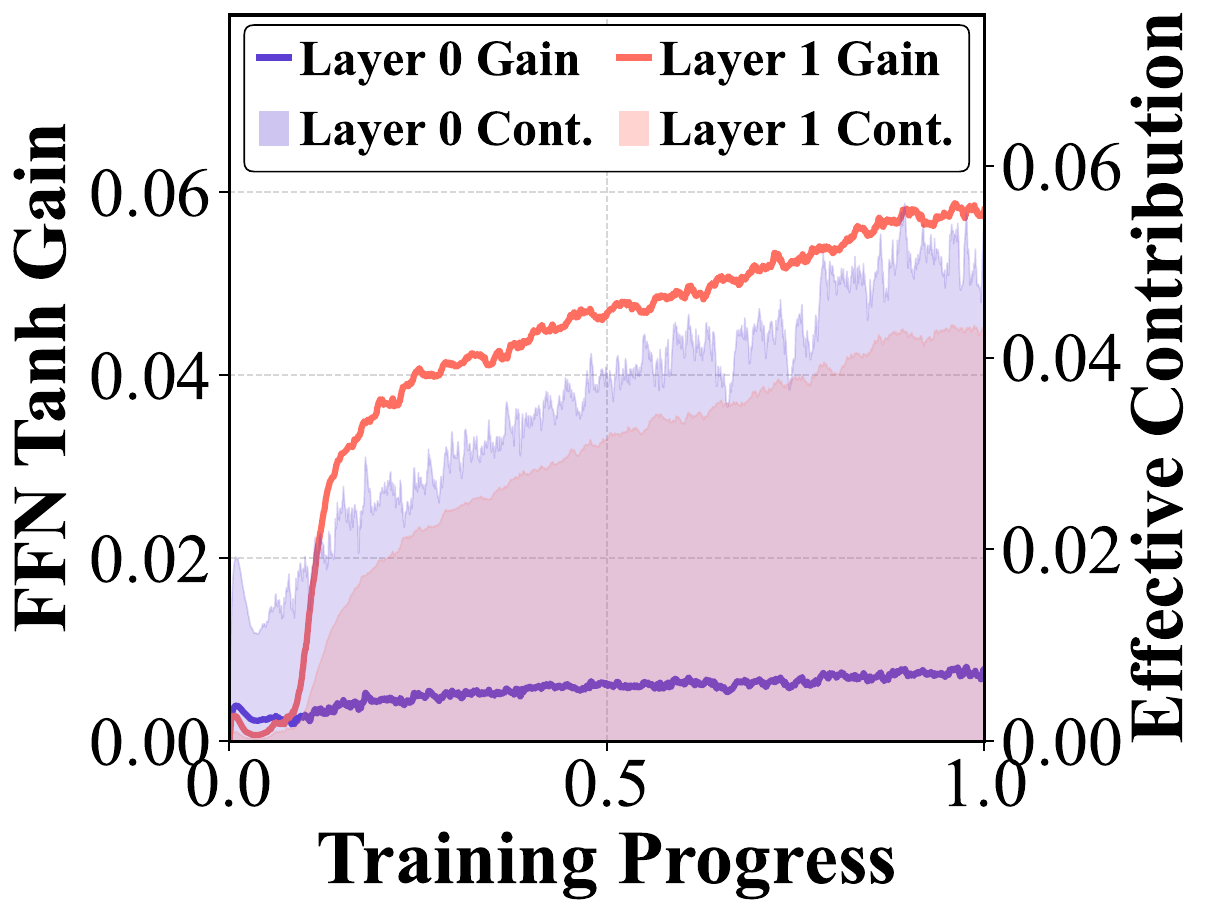}\hfill
\includegraphics[width=.24\textwidth]{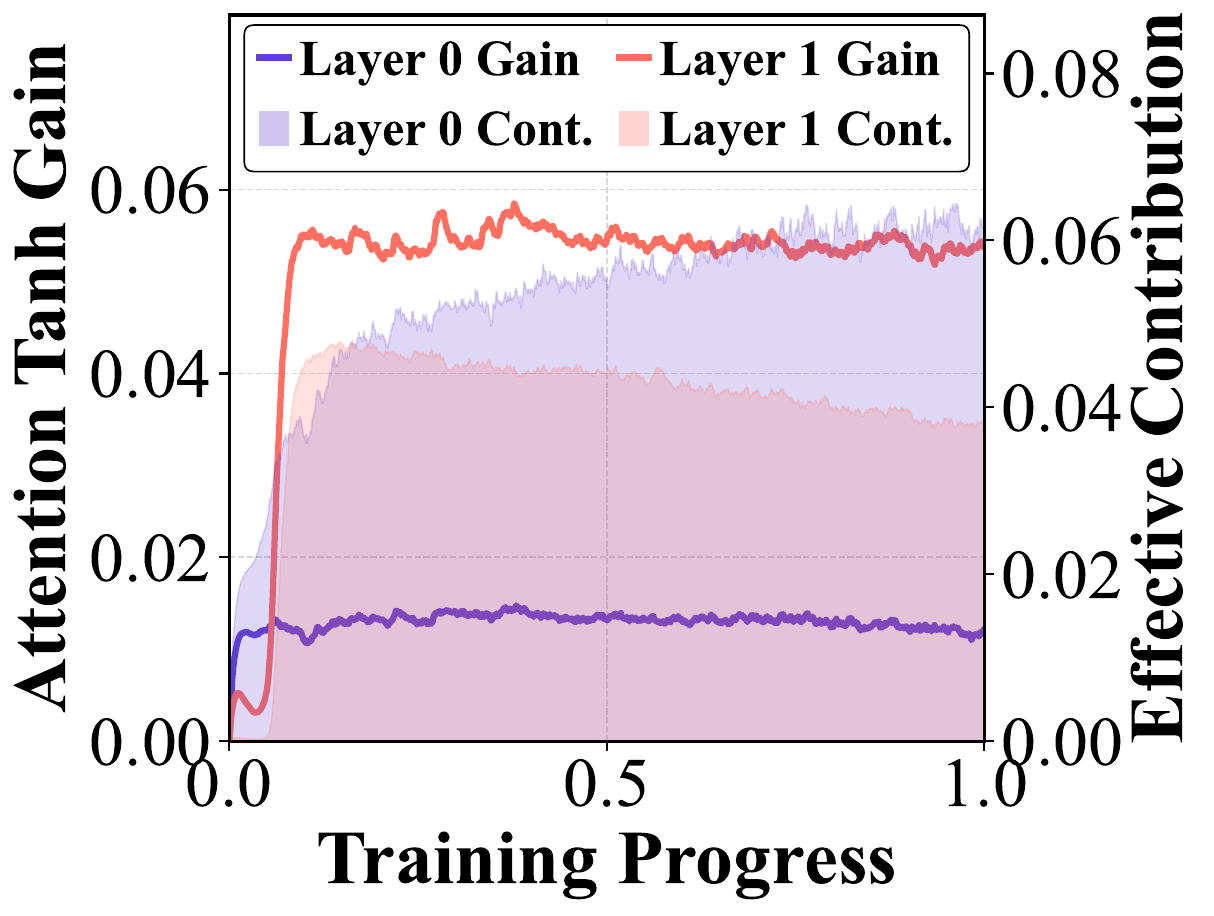}
\caption{Training-time evolution of the global injection gates on the NYC datasets. Layer 0 and Layer 1 denote the two injection modules; ``Cont.'' denotes the
normalized contribution proxy.}
\label{fig:app_gate_evolution}
\end{figure*}

\paragraph{Evidence Utilization.}
As shown in Figure~\ref{fig:evidence_injection}, cross-attention outperforms
prompt concatenation, and learnable gating provides further gains, confirming
the benefit of controlled retrieval from separate memories.
Figure~\ref{fig:app_gate_evolution} shows that the gate gains
$\alpha_{\mathrm{attn}}$ and $\alpha_{\mathrm{ffn}}$, together with their
update contributions, increase from zero and stabilize during training. Details of the Effective Contribution metric and additional insertion-frequency experiments are provided in Appendix~\ref{app:injection_analysis}.

The source ablations show that dynamic memory alone is insufficient and that shuffled context degrades performance, showing that sample-specific context provides complementary information rather than merely additional tokens. Overall, the frozen LLM effectively utilizes both memory sources, with gated injection further improving their integration.

\subsection{Qualitative Case Study}
\begin{figure}[htbp]
    \centering
    \renewcommand{\thesubfigure}{\normalfont\alph{subfigure}}
    \captionsetup[subfigure]{skip=0.1pt}

    {\centering
    \includegraphics[
        width=0.8\columnwidth,
        keepaspectratio
    ]{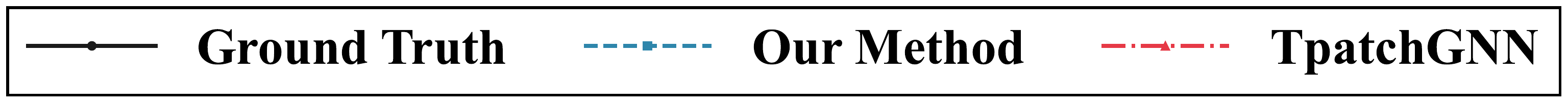}
    \par}

    \vspace{0.1em}

    \begin{subfigure}[t]{\columnwidth}
        \centering
        \includegraphics[
            width=\textwidth,
            keepaspectratio
        ]{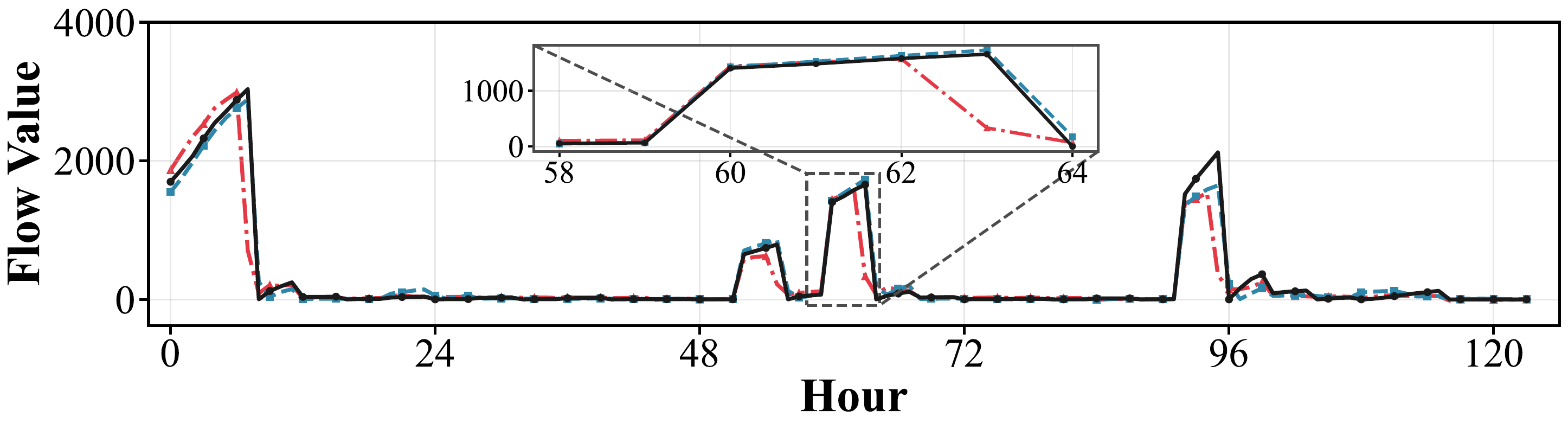}
        \caption{\normalfont Inflow}
        \label{fig:taxi_case_inflow}
    \end{subfigure}

    \begin{subfigure}[t]{\columnwidth}
        \centering
        \includegraphics[
            width=\textwidth,
            keepaspectratio
        ]{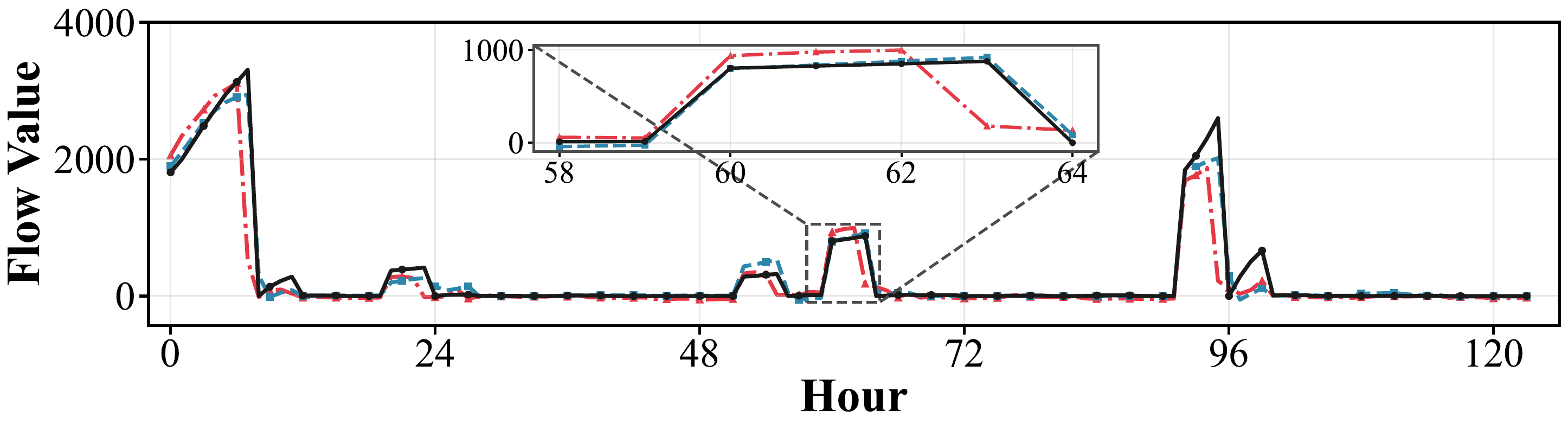}
        \caption{\normalfont Outflow}
        \label{fig:taxi_case_outflow}
    \end{subfigure}

    \vspace{-2.5ex}
    \caption{
    Qualitative comparison on NYCtaxi over a 124-hour evaluation
    period. The enlarged region highlights a representative demand
    peak. LLMODE more closely follows the ground-truth temporal
    patterns than tPatchGNN, particularly around rapid variations.
    }
    \label{fig:nyc-taxi-case-study}
    \vspace{-1ex}
\end{figure}

Figure~\ref{fig:nyc-taxi-case-study} compares LLMODE with tPatchGNN (Zhang et al., 2024), a strong baseline for irregular temporal modeling, on a representative supervised NYCtaxi case. LLMODE follows the ground-truth trajectory more closely and better captures the magnitude and timing of peak variations. This behavior is consistent with its continuous-time modeling, fixed-budget dynamic memory, and gated injection design. For visualization, predictions from multiple samples are concatenated, with details provided in Appendix ~\ref{app:case_study}.

\section{Related Work} \label{sec:6_related}

\noindent{\bf Spatio-temporal Forecasting.}
Deep spatio-temporal forecasting models typically combine temporal modeling
with spatial dependency encoding. Early methods focus on grid-structured data
and integrate recurrent or temporal convolutional networks with convolutional
spatial operators~\cite{dlstm,stresnet}. Graph-based methods instead combine
GNNs with recurrent or convolutional temporal modules~\cite{TGCN,STSGCN,MTGNN,graphwavenet}.
Recent approaches further improve forecasting through adaptive graph learning
and multi-scale or frequency-aware representations~\cite{jin2023transferable,
ye2022learning,UniST,dong2024heterogeneity,wang2024stone,BiTGraph,FourierGNN}.
However, most are developed for supervised and regularly sampled settings,
leaving severe irregularity and zero-shot generalization to unseen regions
relatively underexplored.

\noindent{\bf LLM-based Forecasting.}
LLM-based spatio-temporal forecasting remains relatively limited.
UrbanGPT~\cite{UrbanGPT} represents spatio-temporal signals as soft tokens
and inserts them into the LLM prompt. Related time-series methods either
convert numerical sequences into text~\cite{xue2023promptcast} or align
compact numerical tokens with the LLM representation space through prompting,
projection layers, adapters, or partial fine-tuning~\cite{TEMPO,Fromnews,
IP-LLM,Context-Alignment,LLM4TS,TEST,Time-LLM,TimeCMA,HyperLoad,AutoTimes}.
Similar to multimodal architectures such as Flamingo~\cite{alayrac2022flamingo},
some methods use cross-attention to inject external features. Nevertheless,
robust integration of spatio-temporal evidence under irregular and
asynchronous observations remains underexplored.

\noindent{\bf Irregularly Sampled Time Series Modeling.}
A common strategy regularizes irregular time series through interpolation or
resampling~\cite{lipton2016directly}, which may distort dynamics under sparse
observations or large time gaps~\cite{shukla2021multi}. Direct approaches
include time-aware recurrent networks~\cite{neil2016phased,che2018recurrent,
baytas2017patient,jhin2024addressing}, neural ODEs that evolve latent states
in continuous time~\cite{chen2018neural,rubanova2019latent,GRU-ODE,LG-ODE,
neural_flows,kidger2020neural,dupont2019augmented}, and attention-based models
for missing and nonuniform observations~\cite{tPatchGNN,ViTST}. However, most
studies focus on limited tasks or classification~\cite{zhang2023warpformer,
zhong2025mtm,yehuda2023self}, leaving LLM-based irregular spatio-temporal
forecasting with spatial dependencies underexplored.

\section{Conclusion}
\label{sec:7_discussion}

In this work, we presented LLMODE, a frozen-LLM framework for irregular spatio-temporal forecasting. LLMODE reconstructs continuous-time graph dynamics from asynchronous and partially observed data, distills long latent trajectories into fixed-budget dynamic memory, and injects dynamic and context memories into selected layers of the frozen LLM through dual-source gated cross-attention. Experiments on urban and physical-dynamics benchmarks validate the framework and its key components, demonstrating competitive forecasting performance, improved robustness under sparse observations, strong generalization to unseen regions, and substantially reduced token usage and inference latency compared with existing LLM-based forecasters.


\clearpage
\bibliography{aaai2027}

@article{TGCN,
  title={T-GCN: A temporal graph convolutional network for traffic prediction},
  author={Zhao, Ling and Song, Yujiao and Zhang, Chao and Liu, Yu and Wang, Pu and Lin, Tao and Deng, Min and Li, Haifeng},
  journal={IEEE transactions on intelligent transportation systems},
  volume={21},
  number={9},
  pages={3848--3858},
  year={2019},
  publisher={IEEE}
}

@inproceedings{STSGCN,
  title={Spatial-temporal synchronous graph convolutional networks: A new framework for spatial-temporal network data forecasting},
  author={Song, Chao and Lin, Youfang and Guo, Shengnan and Wan, Huaiyu},
  booktitle={Proceedings of the AAAI conference on artificial intelligence},
  volume={34},
  pages={914--921},
  year={2020}
}

@inproceedings{MTGNN,
  title={Connecting the dots: Multivariate time series forecasting with graph neural networks},
  author={Wu, Zonghan and Pan, Shirui and Long, Guodong and Jiang, Jing and Chang, Xiaojun and Zhang, Chengqi},
  booktitle={Proceedings of the 26th ACM SIGKDD international conference on knowledge discovery \& data mining},
  pages={753--763},
  year={2020}
}

@inproceedings{BiTGraph,
  title={Biased temporal convolution graph network for time series forecasting with missing values},
  author={Chen, Xiaodan and Li, Xiucheng and Liu, Bo and Li, Zhijun},
  booktitle={The Twelfth International Conference on Learning Representations},
  year={2023}
}

@misc{FourierGNN,
      title={FourierGNN: Rethinking Multivariate Time Series Forecasting from a Pure Graph Perspective}, 
      author={Kun Yi and Qi Zhang and Wei Fan and Hui He and Liang Hu and Pengyang Wang and Ning An and Longbing Cao and Zhendong Niu},
      year={2023},
      eprint={2311.06190},
      archivePrefix={arXiv},
      primaryClass={cs.LG},
      url={https://arxiv.org/abs/2311.06190}, 
}

@inproceedings{ISTS-PLM,
  title={Unleashing the power of pre-trained language models for irregularly sampled time series},
  author={Zhang, Weijia and Yin, Chenlong and Liu, Hao and Xiong, Hui},
  booktitle={Proceedings of the 31st ACM SIGKDD Conference on Knowledge Discovery and Data Mining V. 2},
  pages={3831--3842},
  year={2025}
}

@inproceedings{UrbanGPT,
  title={Urbangpt: Spatio-temporal large language models},
  author={Li, Zhonghang and Xia, Lianghao and Tang, Jiabin and Xu, Yong and Shi, Lei and Xia, Long and Yin, Dawei and Huang, Chao},
  booktitle={Proceedings of the 30th ACM SIGKDD Conference on Knowledge Discovery and Data Mining},
  pages={5351--5362},
  year={2024}
}

@article{GPT4TS,
  title={One fits all: Power general time series analysis by pretrained lm},
  author={Zhou, Tian and Niu, Peisong and Sun, Liang and Jin, Rong and others},
  journal={Advances in neural information processing systems},
  volume={36},
  pages={43322--43355},
  year={2023}
}

@article{GRU-ODE,
  title={GRU-ODE-Bayes: Continuous modeling of sporadically-observed time series},
  author={De Brouwer, Edward and Simm, Jaak and Arany, Adam and Moreau, Yves},
  journal={Advances in neural information processing systems},
  volume={32},
  year={2019}
}

@inproceedings{tPatchGNN,
  title={Irregular multivariate time series forecasting: A transformable patching graph neural networks approach},
  author={Zhang, Weijia and Yin, Chenlong and Liu, Hao and Zhou, Xiaofang and Xiong, Hui},
  booktitle={Forty-first International Conference on Machine Learning},
  year={2024}
}

@misc{ViTST,
      title={Time Series as Images: Vision Transformer for Irregularly Sampled Time Series}, 
      author={Zekun Li and Shiyang Li and Xifeng Yan},
      year={2023},
      eprint={2303.12799},
      archivePrefix={arXiv},
      primaryClass={cs.LG},
      url={https://arxiv.org/abs/2303.12799}, 
}

@article{vicuna7b,
  title={Judging llm-as-a-judge with mt-bench and chatbot arena},
  author={Zheng, Lianmin and Chiang, Wei-Lin and Sheng, Ying and Zhuang, Siyuan and Wu, Zhanghao and Zhuang, Yonghao and Lin, Zi and Li, Zhuohan and Li, Dacheng and Xing, Eric and others},
  journal={Advances in neural information processing systems},
  volume={36},
  pages={46595--46623},
  year={2023}
}

@inproceedings{kipf2018neural,
  title={Neural relational inference for interacting systems},
  author={Kipf, Thomas and Fetaya, Ethan and Wang, Kuan-Chieh and Welling, Max and Zemel, Richard},
  booktitle={International conference on machine learning},
  pages={2688--2697},
  year={2018},
  organization={Pmlr}
}

@inproceedings{dlstm,
  title={Deep learning: A generic approach for extreme condition traffic forecasting},
  author={Yu, Rose and Li, Yaguang and Shahabi, Cyrus and Demiryurek, Ugur and Liu, Yan},
  booktitle={Proceedings of the 2017 SIAM international Conference on Data Mining},
  pages={777--785},
  year={2017},
  organization={SIAM}
}

@inproceedings{stresnet,
  title={Deep spatio-temporal residual networks for citywide crowd flows prediction},
  author={Zhang, Junbo and Zheng, Yu and Qi, Dekang},
  booktitle={Proceedings of the AAAI conference on artificial intelligence},
  volume={31},
  number={1},
  year={2017}
}

@article{graphwavenet,
  title={Graph wavenet for deep spatial-temporal graph modeling},
  author={Wu, Zonghan and Pan, Shirui and Long, Guodong and Jiang, Jing and Zhang, Chengqi},
  journal={arXiv preprint arXiv:1906.00121},
  year={2019}
}

@inproceedings{jin2023transferable,
  title={Transferable graph structure learning for graph-based traffic forecasting across cities},
  author={Jin, Yilun and Chen, Kai and Yang, Qiang},
  booktitle={Proceedings of the 29th ACM SIGKDD conference on knowledge discovery and data mining},
  pages={1032--1043},
  year={2023}
}

@inproceedings{ye2022learning,
  title={Learning the evolutionary and multi-scale graph structure for multivariate time series forecasting},
  author={Ye, Junchen and Liu, Zihan and Du, Bowen and Sun, Leilei and Li, Weimiao and Fu, Yanjie and Xiong, Hui},
  booktitle={Proceedings of the 28th ACM SIGKDD conference on knowledge discovery and data mining},
  pages={2296--2306},
  year={2022}
}

@article{che2018recurrent,
  title={Recurrent neural networks for multivariate time series with missing values},
  author={Che, Zhengping and Purushotham, Sanjay and Cho, Kyunghyun and Sontag, David and Liu, Yan},
  journal={Scientific reports},
  volume={8},
  number={1},
  pages={6085},
  year={2018},
  publisher={Nature Publishing Group UK London}
}

@inproceedings{lipton2016directly,
  title={Directly modeling missing data in sequences with rnns: Improved classification of clinical time series},
  author={Lipton, Zachary C and Kale, David and Wetzel, Randall},
  booktitle={Machine learning for healthcare conference},
  pages={253--270},
  year={2016},
  organization={PMLR}
}

@article{shukla2021multi,
  title={Multi-time attention networks for irregularly sampled time series},
  author={Shukla, Satya Narayan and Marlin, Benjamin M},
  journal={arXiv preprint arXiv:2101.10318},
  year={2021}
}

@article{neil2016phased,
  title={Phased lstm: Accelerating recurrent network training for long or event-based sequences},
  author={Neil, Daniel and Pfeiffer, Michael and Liu, Shih-Chii},
  journal={Advances in neural information processing systems},
  volume={29},
  year={2016}
}

@inproceedings{baytas2017patient,
  title={Patient subtyping via time-aware LSTM networks},
  author={Baytas, Inci M and Xiao, Cao and Zhang, Xi and Wang, Fei and Jain, Anil K and Zhou, Jiayu},
  booktitle={Proceedings of the 23rd ACM SIGKDD international conference on knowledge discovery and data mining},
  pages={65--74},
  year={2017}
}

@article{chen2018neural,
  title={Neural ordinary differential equations},
  author={Chen, Ricky TQ and Rubanova, Yulia and Bettencourt, Jesse and Duvenaud, David K},
  journal={Advances in neural information processing systems},
  volume={31},
  year={2018}
}

@article{rubanova2019latent,
  title={Latent ordinary differential equations for irregularly-sampled time series},
  author={Rubanova, Yulia and Chen, Ricky TQ and Duvenaud, David K},
  journal={Advances in neural information processing systems},
  volume={32},
  year={2019}
}

@inproceedings{zhang2023warpformer,
  title={Warpformer: A multi-scale modeling approach for irregular clinical time series},
  author={Zhang, Jiawen and Zheng, Shun and Cao, Wei and Bian, Jiang and Li, Jia},
  booktitle={Proceedings of the 29th ACM SIGKDD Conference on Knowledge Discovery and Data Mining},
  pages={3273--3285},
  year={2023}
}

@inproceedings{yehuda2023self,
  title={Self-supervised classification of clinical multivariate time series using time series dynamics},
  author={Yehuda, Yakir and Freedman, Daniel and Radinsky, Kira},
  booktitle={Proceedings of the 29th ACM SIGKDD Conference on Knowledge Discovery and Data Mining},
  pages={5416--5427},
  year={2023}
}

@inproceedings{zhong2025mtm,
  title={MTM: A Multi-Scale Token Mixing Transformer for Irregular Multivariate Time Series Classification},
  author={Zhong, Shuhan and Zhuo, Weipeng and Song, Sizhe and Li, Guanyao and Yu, Zhongyi and Chan, S-H Gary},
  booktitle={Proceedings of the 31st ACM SIGKDD Conference on Knowledge Discovery and Data Mining V. 2},
  pages={4074--4085},
  year={2025}
}

@article{neural_flows,
  title={Neural flows: Efficient alternative to neural ODEs},
  author={Bilo{\v{s}}, Marin and Sommer, Johanna and Rangapuram, Syama Sundar and Januschowski, Tim and G{\"u}nnemann, Stephan},
  journal={Advances in neural information processing systems},
  volume={34},
  pages={21325--21337},
  year={2021}
}

@article{LG-ODE,
  title={Learning continuous system dynamics from irregularly-sampled partial observations},
  author={Huang, Zijie and Sun, Yizhou and Wang, Wei},
  journal={Advances in Neural Information Processing Systems},
  volume={33},
  pages={16177--16187},
  year={2020}
}

@misc{TEMPO,
  title = {{{TEMPO}}: {{Prompt-based}} Generative Pre-Trained Transformer for Time Series Forecasting},
  shorttitle = {{{TEMPO}}},
  author = {Cao, Defu and Jia, Furong and Arik, Sercan O. and Pfister, Tomas and Zheng, Yixiang and Ye, Wen and Liu, Yan},
  year = 2024,
  month = apr,
  number = {arXiv:2310.04948},
  eprint = {2310.04948},
  primaryclass = {cs},
  publisher = {arXiv},
  doi = {10.48550/arXiv.2310.04948},
  archiveprefix = {arXiv}
}

@article{LLM4TS,
  title = {{{LLM4TS}}: {{Aligning}} Pre-Trained {{LLMs}} as Data-Efficient Time-Series Forecasters},
  shorttitle = {{{LLM4TS}}},
  author = {Chang, Ching and Wang, Wei-Yao and Peng, Wen-Chih and Chen, Tien-Fu},
  year = 2025,
  month = jun,
  journal = {ACM Trans. Intell. Syst. Technol.},
  volume = {16},
  number = {3},
  pages = {1--20},
  issn = {2157-6904, 2157-6912},
  doi = {10.1145/3719207},
  langid = {english}
}

@misc{Context-Alignment,
  title = {Context-{{Alignment}}: {{Activating}} and {{Enhancing LLM Capabilities}} in {{Time Series}}},
  shorttitle = {Context-{{Alignment}}},
  author = {Hu, Yuxiao and Li, Qian and Zhang, Dongxiao and Yan, Jinyue and Chen, Yuntian},
  year = 2025,
  month = apr,
  number = {arXiv:2501.03747},
  eprint = {2501.03747},
  primaryclass = {cs},
  publisher = {arXiv},
  doi = {10.48550/arXiv.2501.03747},
  archiveprefix = {arXiv},
  langid = {american}
}

@misc{HyperLoad,
  title = {{{HyperLoad}}: {{A}} Cross-Modality Enhanced Large Language Model-Based Framework for Green Data Center Cooling Load Prediction},
  shorttitle = {{{HyperLoad}}},
  author = {Jiang, Haoyu and Qu, Boan and Zhu, Junjie and Zeng, Fanjie and Lin, Xiaojie and Zhong, Wei},
  year = 2025,
  month = dec,
  number = {arXiv:2512.19114},
  eprint = {2512.19114},
  primaryclass = {cs},
  publisher = {arXiv},
  doi = {10.48550/arXiv.2512.19114},
  archiveprefix = {arXiv}
}

@article{AutoTimes,
  title = {{{AutoTimes}}: {{Autoregressive}} Time Series Forecasters via Large Language Models},
  shorttitle = {{{AutoTimes}}},
  author = {Liu, Yong and Qin, Guo and Huang, Xiangdong and Wang, Jianmin and Long, Mingsheng},
  year = 2024,
  month = dec,
  journal = {Advances in Neural Information Processing Systems},
  volume = {37},
  pages = {122154--122184},
  doi = {10.52202/079017-3882},
  langid = {english}
}

@inproceedings{TimeCMA,
  title={Timecma: Towards llm-empowered multivariate time series forecasting via cross-modality alignment},
  author={Liu, Chenxi and Xu, Qianxiong and Miao, Hao and Yang, Sun and Zhang, Lingzheng and Long, Cheng and Li, Ziyue and Zhao, Rui},
  booktitle={Proceedings of the AAAI Conference on Artificial Intelligence},
  volume={39},
  number={18},
  pages={18780--18788},
  year={2025}
}

@inproceedings{IP-LLM,
  title = {\${{S}}\textasciicircum 2\${{IP-LLM}}: {{Semantic}} Space Informed Prompt Learning with {{LLM}} for Time Series Forecasting},
  shorttitle = {\${{S}}\textasciicircum 2\${{IP-LLM}}},
  booktitle = {Forty-First {{International Conference}} on {{Machine Learning}}},
  author = {Pan, Zijie and Jiang, Yushan and Garg, Sahil and Schneider, Anderson and Nevmyvaka, Yuriy and Song, Dongjin},
  year = 2024,
  month = jun,
  langid = {english}
}

@misc{TEST,
  title = {{{TEST}}: {{Text}} Prototype Aligned Embedding to Activate {{LLM}}'s Ability for Time Series},
  shorttitle = {{{TEST}}},
  author = {Sun, Chenxi and Li, Hongyan and Li, Yaliang and Hong, Shenda},
  year = 2024,
  month = feb,
  number = {arXiv:2308.08241},
  eprint = {2308.08241},
  primaryclass = {cs},
  publisher = {arXiv},
  doi = {10.48550/arXiv.2308.08241},
  archiveprefix = {arXiv},
  langid = {american}
}

@article{xue2023promptcast,
  title={Promptcast: A new prompt-based learning paradigm for time series forecasting},
  author={Xue, Hao and Salim, Flora D},
  journal={IEEE Transactions on Knowledge and Data Engineering},
  volume={36},
  number={11},
  pages={6851--6864},
  year={2023},
  publisher={IEEE}
}

@article{Fromnews,
  title={From news to forecast: Integrating event analysis in llm-based time series forecasting with reflection},
  author={Wang, Xinlei and Feng, Maike and Qiu, Jing and Gu, Jinjin and Zhao, Junhua},
  journal={Advances in Neural Information Processing Systems},
  volume={37},
  pages={58118--58153},
  year={2024}
}

@article{Time-LLM,
  title={Time-llm: Time series forecasting by reprogramming large language models},
  author={Jin, Ming and Wang, Shiyu and Ma, Lintao and Chu, Zhixuan and Zhang, James Y and Shi, Xiaoming and Chen, Pin-Yu and Liang, Yuxuan and Li, Yuan-Fang and Pan, Shirui and others},
  journal={arXiv preprint arXiv:2310.01728},
  year={2023}
}

@misc{li2024urbangpt,
      title={UrbanGPT: Spatio-Temporal Large Language Models}, 
      author={Zhonghang Li and Lianghao Xia and Jiabin Tang and Yong Xu and Lei Shi and Long Xia and Dawei Yin and Chao Huang},
      year={2024},
      eprint={2403.00813},
      archivePrefix={arXiv},
      primaryClass={cs.CL}
}

@article{alayrac2022flamingo,
  title={Flamingo: a visual language model for few-shot learning},
  author={Alayrac, Jean-Baptiste and Donahue, Jeff and Luc, Pauline and Miech, Antoine and Barr, Iain and Hasson, Yana and Lenc, Karel and Mensch, Arthur and Millican, Katherine and Reynolds, Malcolm and others},
  journal={Advances in neural information processing systems},
  volume={35},
  pages={23716--23736},
  year={2022}
}

@inproceedings{UniST,
  title={Unist: A prompt-empowered universal model for urban spatio-temporal prediction},
  author={Yuan, Yuan and Ding, Jingtao and Feng, Jie and Jin, Depeng and Li, Yong},
  booktitle={Proceedings of the 30th ACM SIGKDD Conference on Knowledge Discovery and Data Mining},
  pages={4095--4106},
  year={2024}
}

@inproceedings{dong2024heterogeneity,
  title={Heterogeneity-informed meta-parameter learning for spatiotemporal time series forecasting},
  author={Dong, Zheng and Jiang, Renhe and Gao, Haotian and Liu, Hangchen and Deng, Jinliang and Wen, Qingsong and Song, Xuan},
  booktitle={Proceedings of the 30th ACM SIGKDD conference on knowledge discovery and data mining},
  pages={631--641},
  year={2024}
}

@inproceedings{wang2024stone,
  title={Stone: A spatio-temporal ood learning framework kills both spatial and temporal shifts},
  author={Wang, Binwu and Ma, Jiaming and Wang, Pengkun and Wang, Xu and Zhang, Yudong and Zhou, Zhengyang and Wang, Yang},
  booktitle={Proceedings of the 30th ACM SIGKDD Conference on Knowledge Discovery and Data Mining},
  pages={2948--2959},
  year={2024}
}

@inproceedings{jhin2024addressing,
  title={Addressing prediction delays in time series forecasting: A continuous gru approach with derivative regularization},
  author={Jhin, Sheo Yon and Kim, Seojin and Park, Noseong},
  booktitle={Proceedings of the 30th ACM SIGKDD Conference on Knowledge Discovery and Data Mining},
  pages={1234--1245},
  year={2024}
}

@article{kidger2020neural,
  title={Neural controlled differential equations for irregular time series},
  author={Kidger, Patrick and Morrill, James and Foster, James and Lyons, Terry},
  journal={Advances in neural information processing systems},
  volume={33},
  pages={6696--6707},
  year={2020}
}

@article{dupont2019augmented,
  title={Augmented neural odes},
  author={Dupont, Emilien and Doucet, Arnaud and Teh, Yee Whye},
  journal={Advances in neural information processing systems},
  volume={32},
  year={2019}
}
\clearpage
\appendix
\section{Appendix}
\label{sec:A_supplementary_experiments}

\subsection{Dataset Details}
\label{app:datasets_construction}

\paragraph{Urban Datasets (NYC)}

We evaluate on three New York City spatio-temporal datasets: \textbf{NYCtaxi}, \textbf{NYCbike}, and \textbf{NYCcrime}, which capture taxi flows, bike-share flows, and crime incidents, respectively.
Following common practice in urban computing, we partition the city into grid-like regions using latitude and longitude coordinates.
NYCtaxi contains 263 regions at approximately $3\,\mathrm{km}\times 3\,\mathrm{km}$ resolution, while NYCbike and NYCcrime contain 2162 regions at approximately $1\,\mathrm{km}\times 1\,\mathrm{km}$ resolution.
To better reflect real-world urban data characteristics, we formulate all three datasets as \emph{irregularly sampled} sequences.

\subparagraph{Event-Level Source Records.}
The original NYC datasets are collected in an event-level format, where each row records an individual event rather than a value aggregated over a predefined temporal grid.
For example, in NYCtaxi, each row corresponds to a single taxi trip.
Table~\ref{tab:event_level_example} shows one representative record.
The \texttt{vendor\_id} identifies the data provider, while \texttt{pickup\_datetime} and \texttt{dropoff\_datetime} indicate the start and end times of the trip.
The \texttt{passenger\_count} records the number of passengers, and the pickup and drop-off longitude--latitude pairs specify the origin and destination of the event.
Therefore, the raw dataset consists of asynchronously occurring trips with event-specific timestamps and spatial locations, rather than regularly sampled regional time series.

\begin{table*}[t]
\centering
\scriptsize
\setlength{\tabcolsep}{3.0pt}
\caption{An example of an event-level taxi-trip record.}
\label{tab:event_level_example}
\resizebox{\textwidth}{!}{
\begin{tabular}{@{}cccccccc@{}}
\toprule
\texttt{vendor\_id}
& \texttt{pickup\_datetime}
& \texttt{dropoff\_datetime}
& \texttt{passenger\_count}
& \texttt{pickup\_longitude}
& \texttt{pickup\_latitude}
& \texttt{dropoff\_longitude}
& \texttt{dropoff\_latitude} \\
\midrule
1
& 2016-03-14 17:24:55
& 2016-03-14 17:32:30
& 1
& -73.982154846191406
& 40.767936706542969
& -73.964630126953125
& 40.765602111816406 \\
\bottomrule
\end{tabular}
}
\end{table*}

To construct the spatio-temporal forecasting benchmarks, we first assign individual events to spatial regions according to their coordinates and then aggregate them into region-level signals.
For NYCtaxi and NYCbike, pickup and drop-off events are converted into regional inflow and outflow counts, while NYCcrime events are aggregated by region and crime category.
In this way, the original event stream is transformed into graph-structured node observations, where each node corresponds to an urban region.

\subparagraph{Irregular Sampling Construction.}
Unlike standard time-series preprocessing that places all observations on a uniformly spaced temporal grid, we retain non-uniform observation intervals when constructing the input sequences.
Consequently, consecutive observations can be separated by different elapsed times, and different spatial nodes may contain asynchronous or incomplete histories.
For NYCtaxi and NYCbike, the resulting sampling intervals range from 10 to 360 minutes, with an average interval of approximately 60 minutes.
For NYCcrime, the intervals range from 1 to 16 days, with an average interval of approximately 4 days.
This construction preserves the event-driven and irregular characteristics of the original data and provides a realistic setting for evaluating forecasting models under non-uniform timestamps and missing observations.

\subparagraph{Task Setup.}
We define the observations and forecasting horizons as follows.
For taxi and bike datasets, the observation at time $t$ is defined as the cumulative flow from 00:00 of the same day up to time $t$.
For crime data, we use a sliding time window to compute a sequence of cumulative crime counts within the window as observations.
We adopt forecasting horizons of using the past 24 hours to predict the next 12 hours for taxi/bike datasets, and using the past 96 days to predict the next 48 days for crime data.

\subparagraph{Training and Evaluation Protocols.}
Following the experimental protocol of UrbanGPT~\cite{UrbanGPT}, during the instruction-tuning stage, we randomly select 80 regions from the three NYC datasets as training data.
The region indices are kept consistent for NYCbike and NYCcrime datasets.
We use the following time spans for instruction tuning: for \textbf{NYCtaxi}, Jan.~1, 2017 to Mar.~31, 2017; for \textbf{NYCbike}, Apr.~1, 2017 to Jun.~30, 2017; and for \textbf{NYCcrime}, Jan.~1, 2016 to Dec.~31, 2018.
For pretraining the spatio-temporal dependency encoder and training baseline models, we use the same training data and set the maximum number of epochs to 100.

In the testing stage, we conduct two evaluation settings:
\begin{enumerate}
    \item \textbf{Supervised prediction:} We evaluate the model on a long continuous interval using all data from Dec.~2021 for NYCtaxi and NYCbike, and all data from the entire year of 2021 for NYCcrime.
    
    \item \textbf{Zero-shot prediction:} We use an additional 80 regions from the NYC datasets as unseen test regions. For NYCtaxi and NYCbike, we use the first two weeks of data in 2020 for testing. For NYCcrime, we use the entire year of 2020 for testing.
\end{enumerate}

\subparagraph{Normalization.}
We normalize each NYC dataset independently using statistics computed from its \emph{training split} only, and apply the same transformation to validation/test data to avoid information leakage.
Specifically, for each variable we perform z-score normalization:
\begin{equation}
\tilde{x} =
\frac{x-\mu_{\mathrm{train}}}
{\sigma_{\mathrm{train}}},
\end{equation}
where $\mu_{\mathrm{train}}$ and $\sigma_{\mathrm{train}}$ denote the mean and standard deviation estimated on the training split.

\paragraph{Physics-Based Simulation Datasets}

We additionally evaluate our model on two physics-based simulated datasets, namely \textbf{Springs} and \textbf{Charged}, following Kipf et al.~\cite{kipf2018neural}.
Each sample contains five interacting particles moving in a 2D box without external forces, where particles may collide with the box boundaries but do not experience friction or damping.
Using a standard physics simulator, we generate trajectories of 60{,}000
integration steps for Springs and 12{,}000 steps for Charged.

\subparagraph{Irregular Sampling Construction.}
To simulate realistic sensing scenarios with limited and asynchronous observations, we construct irregularly sampled partial observations as follows.
First, we reduce the temporal resolution by subsampling the simulated trajectories using different downsampling factors: 500 for the Springs system and 100 for the charged system, reflecting their different characteristic timescales.
Then, for each particle independently, we sample the number of observations $n \sim \mathcal{U}(20,28)$ and select $n$ observation timestamps uniformly from the subsampled time grid.
This procedure creates asynchronous observations where different particles are observed at different times, mimicking real-world sensor networks.

\subparagraph{Task Setup.}
We formulate the forecasting task on the subsampled trajectories.
For both systems, the first 60 sampled time steps define the history window, from which $n\sim\mathcal{U}(20,28)$ timestamps are retained
as irregular observations. The subsequent 60 sampled time steps are used as prediction targets.
Due to different downsampling factors, this corresponds to different numbers of original integration steps: 30,000 steps for the Springs system (downsampling factor 500) and 6,000 steps for the charged particles system (downsampling factor 100).
This extrapolation setting evaluates the model's ability to forecast long-term dynamics beyond the observation window.

\subparagraph{Normalization.}
We compute the min--max statistics \emph{only} on the training split and apply the same transformation to validation/test data to avoid information leakage.
Specifically, we linearly rescale each state variable to the range $[-1,1]$:
\begin{equation}
\tilde{x}
=
2 \cdot
\frac{x-x_{\min}^{\mathrm{train}}}
{x_{\max}^{\mathrm{train}}-x_{\min}^{\mathrm{train}}}
-1,
\end{equation}
where $x_{\min}^{\mathrm{train}}$ and $x_{\max}^{\mathrm{train}}$ denote the minimum and maximum values of the corresponding state variable over the training split.

\subsection{Training Configuration}
\label{app:training_configuration}

LLMODE is trained with a two-stage strategy to separately optimize the
continuous dynamic representation learning and the frozen-LLM forecasting
objective.

\noindent\underline{\it {Stage 1: Training the Graph-Aware ODE Encoder.}}
In the first stage, we pretrain the graph-aware ODE encoder using the GRU-ODE-Bayes parameterization~\cite{GRU-ODE}, which
jointly considers the reconstruction likelihood of observed features and the
Bayesian posterior consistency regularization:
\begin{equation}
\mathcal{L}_{\mathrm{ODE}}
=
\sum_k
\left(
\mathcal{L}_{\mathrm{pre}}^{(k)}
+
\beta_{\mathrm{post}}
\mathcal{L}_{\mathrm{post}}^{(k)}
\right),
\end{equation}
where $\mathcal{L}_{\mathrm{pre}}^{(k)}$ and
$\mathcal{L}_{\mathrm{post}}^{(k)}$ denote the pre-update reconstruction loss
and the post-update posterior regularization loss at observation step $k$,
respectively, and $\beta_{\mathrm{post}}$ controls the strength of the
posterior regularization. Both losses are computed only on observed
dimensions to accommodate feature-wise missing observations.

After this stage, the graph-aware ODE encoder generates continuous-time latent trajectories 
$\mathbf{H}_{\mathrm{dyn}}$ from irregular observations, which are subsequently distilled into dynamic memory tokens for the subsequent forecasting stage.

\noindent\underline{\it {Stage 2: Training the Frozen-LLM Forecasting Framework.}}
In the second stage, we optimize the forecasting framework while keeping the
pretrained LLM backbone and the graph-aware ODE encoder frozen.
Specifically, the trainable parameters include the dynamic memory resampler,
the language encoder and context-memory resampler, dual-source gated cross-attention
modules, and the forecasting regression head. The
parameters of the LLM backbone and the graph-aware ODE encoder are not
updated during training.

The final objective minimizes the prediction error between the forecasts and
ground-truth states at all query timestamps:
\begin{equation}
\mathcal{L}_{\mathrm{forecast}}
=
\frac{1}{\sum_v |Q_v|}
\sum_v
\sum_{t\in Q_v}
\left\|
\hat{\mathbf y}_v(t)-\mathbf y_v(t)
\right\|_2^2 ,
\end{equation}
where $Q_v$ denotes the query timestamps of node $v$,
$\widehat{\mathbf{y}}_v(t),\mathbf{y}_v(t)
\in\mathbb{R}^{D}$
denote the predicted and ground-truth multivariate target
vectors, respectively.

\subsection{Implementation Configuration}
\label{app:implementation}

\paragraph{Model configuration.}
The latent dimension of the graph-aware ODE encoder is set to
$d_{\mathrm{ode}}=50$. The ODE states and text-encoder outputs are
separately projected into the memory space of dimension
$d_{\mathrm{mem}}$ before dynamic- and context-memory resampling,
respectively. For Vicuna-7B, the LLM hidden dimension is
$d_{\mathrm{LLM}}=4096$. Both memory streams are compressed
using a two-layer Perceiver Resampler with eight attention heads
and a per-head dimension of 64.
The fixed token budget is set to $K=5$ for both the dynamic and
context memories. The strength coefficient of the dynamics-aware attention bias is
set to $\beta=0.1$, and the mixing coefficient $\lambda$ is set
to 0.5 to balance the contributions of the two memories. For
Vicuna-7B, the dual-source gated cross-attention modules are
inserted after Transformer Layers 1 and 16. The maximum prompt
length is set to 2048 tokens.

\paragraph{Optimization and hardware.}
The trainable modules are optimized using AdamW with a learning
rate of $2\times10^{-4}$, cosine learning-rate scheduling, and a
warmup ratio of 0.03. We use a per-device batch size of 32 and
apply early stopping according to the validation loss with a
patience of 10 epochs. All experiments are repeated using five
random seeds, i.e., 42, 43, 44, 45, and 46, and the average results
are reported. All results are obtained from the best validation
checkpoint. Training is performed in BF16 precision on NVIDIA
A100 GPUs.

\subsection{GRU-ODE Parameterization and Continuous Propagation}
\label{app:GRU-ODE parameterization}

For each node $v$, the latent state evolves according to the Neural ODE in
Eq.~\ref{eq:latent_ode}. Following the GRU-ODE formulation~\cite{GRU-ODE}, we
instantiate the vector field as
\begin{equation}
\frac{d\mathbf{h}_v(t)}{dt}
=f_{\theta}(\mathbf{h}_v(t))
=
\bigl(1-\mathbf{z}_v(t)\bigr)\odot
\bigl(\mathbf{g}_v(t)-\mathbf{h}_v(t)\bigr),
\label{eq:app_gru_ode}
\end{equation}
where $\mathbf{z}_v(t)$ is the update gate, $\mathbf{g}_v(t)$ is the candidate
state, and $\odot$ denotes element-wise multiplication. Both quantities are
computed from the current latent state through learnable projections included
in $\theta$. This parameterization gives the continuous vector field a gated
update structure analogous to a discrete GRU while preserving continuous-time
state evolution.

Let $t_{i-1}$ and $t_i$ be two consecutive observation times of node $v$. The
pre-update state at $t_i$ is obtained by numerical integration:
\begin{equation}
\mathbf{h}_v(t_i^-)
=
\operatorname{ODESolve}\!\left(
 f_{\theta},
 \bar{\mathbf{h}}_v(t_{i-1}^{+}),
 [t_{i-1},t_i]
\right).
\label{eq:app_ode_solve}
\end{equation}
Here, $\operatorname{ODESolve}(\cdot)$ denotes a numerical ODE solver that
integrates the vector field $f_{\theta}$ from the initial state
$\bar{\mathbf{h}}_v(t_{i-1}^{+})$ over the interval $[t_{i-1},t_i]$,
yielding the propagated state $\mathbf{h}_v(t_i^-)$ immediately before the
observation at $t_i$ is incorporated.
We use the Euler method as the default solver due to its simplicity and
computational efficiency. Because the integration interval explicitly uses $t_i-t_{i-1}$, the encoder
naturally handles nonuniform sampling gaps and can be queried at arbitrary
intermediate timestamps. The graph-corrected state
$\bar{\mathbf{h}}_v(t_{i-1}^{+})$ is used as the initial condition for the
next interval.

\subsection{Observation Encoding and State Correction}
\label{app:observation encoding}

At time $t_i$, $\mathbf{h}_v(t_i^-)$ represents the propagated latent state
before the new observation $\mathbf{x}_{v,i}$ is incorporated.
The role of $\Phi$ is to convert this observation into a correction signal
for updating the propagated state.
To do so, $\Phi$ first estimates what the observation should be according
to the current latent state, and then compares this estimate with the
actually observed value.

Under a Gaussian observation model, we use a learnable observation head
$\Gamma_{\omega}$ to predict the feature-wise mean and scale from the
propagated latent state:
\begin{equation}
\left[
\boldsymbol{\mu}_{v,i};
\boldsymbol{\sigma}_{v,i}
\right]
=
\Gamma_{\omega}\!\left(
\mathbf{h}_v(t_i^-)
\right),
\qquad
\Gamma_{\omega}: \mathbb{R}^{d_{\mathrm{ode}}} \rightarrow \mathbb{R}^{2D},
\label{eq:app_obs_param}
\end{equation}
where $\Gamma_{\omega}$ is a lightweight neural network parameterized by
$\omega$. It maps the $d_{\mathrm{ode}}$-dimensional latent state to the predicted mean
$\boldsymbol{\mu}_{v,i}\in\mathbb{R}^{D}$ and scale
$\boldsymbol{\sigma}_{v,i}\in\mathbb{R}^{D}$ of the $D$ input features.
The predicted mean represents the observation expected from the current
latent state, while the predicted scale characterizes the corresponding
feature-wise uncertainty.

For the $r$-th feature, we compute the normalized discrepancy between the
new observation and its predicted value:
\begin{equation}
\delta_{v,i,r}
=
\frac{x_{v,i,r}-\mu_{v,i,r}}
{\sigma_{v,i,r}+\epsilon},
\qquad r=1,\ldots,D,
\label{eq:app_obs_discrepancy}
\end{equation}
where $\epsilon$ is a small constant for numerical stability.
A large magnitude of $\delta_{v,i,r}$ indicates that the newly observed
value differs substantially from the value expected by the propagated
latent state.

We then combine the predicted observation, the actual observation, and
their discrepancy into a feature-wise descriptor:
\begin{equation}
\mathbf{s}_{v,i,r}
=
\left[
\mu_{v,i,r};
\sigma_{v,i,r};
x_{v,i,r};
\delta_{v,i,r}
\right].
\label{eq:app_obs_descriptor}
\end{equation}
Each descriptor is mapped to a feature embedding:
\begin{equation}
\mathbf{u}_{v,i,r}
=
\rho\!\left(
\mathbf{A}_{r}\mathbf{s}_{v,i,r}
+
\mathbf{b}_{r}
\right),
\label{eq:app_feature_embedding}
\end{equation}
where $\mathbf{A}_{r}$ and $\mathbf{b}_{r}$ are learnable parameters for
the $r$-th feature, and $\rho(\cdot)$ is a nonlinear activation function.

Finally, the feature embeddings are concatenated to obtain the complete
observation encoding:
\begin{equation}
\Phi\!\left(
\mathbf{x}_{v,i},
\mathbf{h}_v(t_i^-)
\right)
=
\left[
\mathbf{u}_{v,i,1};
\ldots;
\mathbf{u}_{v,i,D}
\right].
\label{eq:app_obs_encoder}
\end{equation}
Therefore, $\Phi$ summarizes what the current latent state predicts, what
is actually observed, and how much the two differ.
The correction GRU combines this observation information with
$\mathbf{h}_v(t_i^-)$ to obtain the corrected state
$\mathbf{h}_v(t_i^+)$ in Eq.~\ref{eq:observation_jump}.
This observation-driven correction anchors the continuously propagated
state to newly arrived measurements and reduces accumulated propagation
errors.

\subsection{Graph-Aware State Correction}
\label{app:graph-aware correction}

After the observation-driven jump, all node states at the current event time
are aligned by propagating nodes without new observations to the same
timestamp. We then use the graph structure to refine each node state with
information from its neighboring nodes.

For each edge $(u,v)\in\mathcal{E}$, we compute a relative message from node
$u$ to node $v$:
\begin{equation}
\begin{aligned}
\mathbf{m}_{u,v}
={}&
\psi\!\left(
\mathbf{h}_u(t_i^+)-\mathbf{h}_v(t_i^+)
\right) \\
&-
\psi\!\left(
\mathbf{h}_v(t_i^+)-\mathbf{h}_u(t_i^+)
\right),
\end{aligned}
\label{eq:app_graph_message}
\end{equation}
where $\psi(\cdot)$ is a lightweight MLP. The messages from all neighboring
nodes are aggregated as
\begin{equation}
\mathbf{a}_v
=
\sum_{u\in\mathcal{N}(v)}
\mathbf{m}_{u,v},
\label{eq:app_graph_aggregation}
\end{equation}
where $\mathcal{N}(v)$ denotes the set of neighboring nodes of node $v$, and
$\mathbf{a}_v$ summarizes the structural information received by node $v$.

A node-specific gate is then computed to control how much neighboring
information is incorporated:
\begin{equation}
\boldsymbol{\eta}_v
=
\sigma\!\left(
\mathbf{W}_{\eta}
[\mathbf{h}_v(t_i^+);\mathbf{a}_v]
+
\mathbf{b}_{\eta}
\right),
\label{eq:app_graph_gate}
\end{equation}
where $[\cdot;\cdot]$ denotes concatenation and $\sigma(\cdot)$ is the
sigmoid function. The graph-aware state is obtained through a residual
correction:
\begin{equation}
\bar{\mathbf{h}}_v(t_i^+)
=
\mathbf{h}_v(t_i^+)
+
\gamma\,
\boldsymbol{\eta}_v
\odot
\mathbf{a}_v,
\label{eq:app_graph_correction}
\end{equation}
where $\odot$ denotes element-wise multiplication and $\gamma$ controls the
overall correction strength.

The relative messages characterize the state differences between connected
nodes, while the gate adaptively determines how strongly the aggregated
neighboring information modifies the current node state. The resulting
$\bar{\mathbf{h}}_v(t_i^+)$ is used as the initial state for the next
continuous propagation interval and is also recorded to construct the latent
trajectory $\mathbf{H}_{\mathrm{dyn}}$.

\subsection{Perceiver-Based Fixed-Budget Resampling}
\label{app:detailed resampling}

For clarity, we omit the node index $v$ in this subsection and describe the resampling for a single-node trajectory. The Perceiver Resampler maintains $K$ learnable latent queries
$\{\mathbf{q}_k\}_{k=1}^{K}$, where
$\mathbf{q}_k\in\mathbb{R}^{d_{\mathrm{mem}}}$.
For each latent state $\mathbf{h}(t_i)$, the corresponding key and value
vectors are computed as
\begin{equation}
\mathbf{k}_i
=
\mathbf{W}_{K}\mathbf{h}(t_i),
\qquad
\mathbf{v}_i
=
\mathbf{W}_{V}\mathbf{h}(t_i),
\label{eq:app_resampler_kv}
\end{equation}
where
$\mathbf{W}_{K},\mathbf{W}_{V}
\in
\mathbb{R}^{d_{\mathrm{mem}}\times d_{\mathrm{ode}}}$
are learnable mappings, and
$\mathbf{k}_i,\mathbf{v}_i
\in
\mathbb{R}^{d_{\mathrm{mem}}}$.

To encourage the resampler to preserve states from rapidly changing
intervals, we directly introduce the magnitude of the instantaneous latent
change into the attention logits:
\begin{equation}
\alpha_{k,i}
=
\operatorname{softmax}_{i}\!\left(
\frac{\mathbf{q}_{k}^{\top}\mathbf{k}_{i}}
{\sqrt{d_{\mathrm{mem}}}}
+
\beta\,
\operatorname{Norm}_{1:L}\!\left(
\left\|
f_{\theta}\!\left(\mathbf{h}(t_i)\right)
\right\|_2
\right)
\right).
\label{eq:app_dynamics_attention}
\end{equation}
Here, $\alpha_{k,i}$ is the attention weight from the $k$-th latent query to
the state at timestamp $t_i$;
$f_{\theta}(\mathbf{h}(t_i))=\dot{\mathbf{h}}(t_i)$ is the instantaneous
latent change determined by the ODE vector field;
$\operatorname{Norm}_{1:L}(\cdot)$ denotes normalization over the $L$
timestamps of the current trajectory; and $\beta$ controls the strength of
the dynamics-aware bias.

The representation produced by the $k$-th latent query is obtained by
aggregating the value vectors:
\begin{equation}
\mathbf{z}_k
=
\sum_{i=1}^{L}
\alpha_{k,i}\mathbf{v}_i,
\qquad
\mathbf{z}_k\in\mathbb{R}^{d_{\mathrm{mem}}}.
\label{eq:app_resampled_representation}
\end{equation}
Because $K$ is fixed independently of the original trajectory length $L$,
the resampler converts trajectories of different lengths into a fixed number
of dynamic memory tokens.

\subsection{Explicit Time Labeling}
\label{app:time-labeling}

Each resampled vector $\mathbf{z}_k$ aggregates information from
multiple timestamps. To indicate the temporal location of the selected
dynamic content, we compute an attention-weighted center time using the same
attention weights:
\begin{equation}
t_k^{\mathrm{center}}
=
\sum_{i=1}^{L}
\alpha_{k,i}t_i.
\label{eq:app_token_center_time}
\end{equation}
The center time represents the temporal position around which the information
selected by the $k$-th latent query is concentrated.

We first encode $t_k^{\mathrm{center}}$ using multi-frequency sinusoidal
features:
\begin{equation}
\boldsymbol{\phi}(t)
=
[
\sin(\omega_1t),\cos(\omega_1t),
\ldots,
\sin(\omega_{M_{\tau}}t),\cos(\omega_{M_{\tau}}t)
].
\label{eq:app_time_features}
\end{equation}
The sinusoidal features are then transformed into a $d_{\mathrm{mem}}$-dimensional time
representation:
\begin{equation}
\boldsymbol{\tau}_k
=
\Gamma_{\tau}\!\left(
\boldsymbol{\phi}(t_k^{\mathrm{center}})
\right)
\in\mathbb{R}^{d_{\mathrm{mem}}},
\label{eq:app_time_embedding}
\end{equation}
where $\Gamma_{\tau}$ is a lightweight learnable network for time encoding.

The time representation is fused with the resampled dynamic content through
element-wise addition:
\begin{equation}
\tilde{\mathbf{z}}_k
=
\mathbf{z}_k
+
\boldsymbol{\tau}_k,
\qquad
\tilde{\mathbf{z}}_k
\in\mathbb{R}^{d_{\mathrm{mem}}}.
\label{eq:app_time_labeled_representation}
\end{equation}
Finally, the $K$ time-aware dynamic memory tokens are stacked as
\begin{equation}
\tilde{\mathbf{Z}}
=
[\tilde{\mathbf{z}}_1;\ldots;\tilde{\mathbf{z}}_K]
\in\mathbb{R}^{K\times d_{\mathrm{mem}}}.
\label{eq:app_dynamics_memory}
\end{equation}
The resulting $\tilde{\mathbf{Z}}$ serves as the dynamic memory for subsequent alignment and interaction with the query-slot hidden states in the frozen LLM.

\subsection{Construction of Context Memory}
\label{app:context_memory}

\noindent\underline{\it {Descriptor format.}}
We first summarize the historical observations into segment-wise statistics
and serialize them into a compact, structured textual description of the form
\begin{quote}\small
\texttt{[SEG\_STATS S=$S$] seg1: ... seg$S$: ... max\_gap=... [/SEG\_STATS]}
\end{quote}
where $S$ is the number of segments used to partition the history window.

\noindent\underline{\it {Meaning of fields.}}
For each segment \texttt{seg$j$}, we report:
(i) \texttt{*\_mean}: the mean value of the corresponding variable within the
segment (computed over observed points);
(ii) \texttt{[max: *, min: *]}: the maximum and minimum values within the
segment (over observed points);
(iii) \texttt{obs\_rate}: the observation rate in the segment, i.e., the
fraction of timestamps in the segment that are observed.
In addition, \texttt{max\_gap} denotes the time gap between the last observed
timestamp in the history window and the first query time to be predicted.

After obtaining the textual description, we encode it into context token representations, i.e., contextualized
token representations using a pretrained text encoder:
\begin{equation}
\mathbf E_P
=
[
\mathbf e_1;\ldots;\mathbf e_{L_p'}
]
\in\mathbb{R}^{L_p'\times d_{\mathrm{mem}}},
\end{equation}
where $L_p'$ is the number of contextualized token representations,
and $d_{\mathrm{mem}}$ is the memory-token dimension after projecting
the text-encoder outputs. 
We then apply the context-memory resampler, which uses the same latent-query resampling mechanism introduced in Sec.~\ref{sec:token_compression}, without the dynamics-aware attention bias
and explicit time labeling, to compress the variable-length token representations 
into a fixed number of context memory tokens:
\begin{equation}
\mathbf P
=
[
\mathbf p_1;\ldots;\mathbf p_K
]
\in\mathbb{R}^{K\times d_{\mathrm{mem}}},
\end{equation}
where $K$ is the fixed number of context memory tokens.

\noindent\underline{\it {Example: Springs.}}~\\
{\small
\noindent
\texttt{[SEG\_STATS S=4]}\\
\texttt{seg1: pos\_mean=[x: -0.05, y: -0.10], x=[max: 0.11, min: -0.16], y=[max: 0.04, min: -0.19], obs\_rate=60.00\%}\\
\texttt{seg2: pos\_mean=[x: 0.16, y: -0.72], x=[max: 0.35, min: 0.05], y=[max: -0.47, min: -0.85], obs\_rate=20.00\%}\\
\texttt{seg3: pos\_mean=[x: 0.31, y: -0.74], x=[max: 0.52, min: 0.06], y=[max: -0.67, min: -0.83], obs\_rate=20.00\%}\\
\texttt{seg4: pos\_mean=[x: 0.33, y: -0.65], x=[max: 0.42, min: 0.24], y=[max: -0.35, min: -0.98], obs\_rate=40.00\% max\_gap=0.00}\\
\texttt{[/SEG\_STATS]}
}

Here, \texttt{pos\_mean} summarizes the average 2D position $(x,y)$ in each
segment, while \texttt{x=[max,min]} and \texttt{y=[max,min]} capture the range
of motion along each axis. The \texttt{obs\_rate} indicates how densely the
trajectory is observed within the segment, and \texttt{max\_gap} measures the
gap from the last observation to the first query time.

\noindent\underline{\it {Example: NYCtaxi.}}~\\
{\small
\noindent
\texttt{[SEG\_STATS S=4]}\\
\texttt{seg1: inflow\_mean=-0.34, outflow\_mean=-0.26, inflow=[max: -0.34, min: -0.34], outflow=[max: -0.26, min: -0.26], obs\_rate=83.33\%}\\
\texttt{seg2: inflow\_mean=-0.29, outflow\_mean=-0.26, inflow=[max: -0.27, min: -0.32], outflow=[max: -0.25, min: -0.26], obs\_rate=100.00\%}\\
\texttt{seg3: inflow\_mean=-0.23, outflow\_mean=-0.25, inflow=[max: -0.18, min: -0.26], outflow=[max: -0.25, min: -0.25], obs\_rate=50.00\%}\\
\texttt{seg4: inflow\_mean=-0.08, outflow\_mean=-0.24, inflow=[max: -0.07, min: -0.11], outflow=[max: -0.24, min: -0.24], obs\_rate=83.33\% max\_gap=5.00}\\
\texttt{[/SEG\_STATS]}}

This snippet summarizes the segment-wise statistics of \texttt{inflow} and
\texttt{outflow} for a region. The \texttt{*\_mean} fields describe the
average level within each segment, the \texttt{[max,min]} fields capture
within-segment variation, \texttt{obs\_rate} reflects missingness, and
\texttt{max\_gap} indicates the forecast gap from the last observed record to
the first query slot.

\subsection{Prompt Examples}
\label{app:prompt_details}

Below, we provide representative prompt examples for the urban and
physical-dynamics datasets.

\begin{tcolorbox}[promptbox, title={NYCtaxi region prompt.}]
\small
\textbf{Task}: Predict future inflow and outflow at the query slots given historical observations over NYC regions, where each node corresponds to a region described by its POI profile.\\
\textbf{Region}: Bronx area with dominant POIs: Cultural Facility,
Residential, and Education Facility.\\
\textbf{Query-Slots}: <$q_1$> <$q_2$> $\ldots$ <$q_{L_q}$>.
\end{tcolorbox}

\begin{tcolorbox}[promptbox, title={Charged interacting-particle prompt.}]
\small
\textbf{Task}: Given irregularly sampled observations of the particle system, predict each particle's future 2D position $(x,y)$ at the query slots.\\
\textbf{System}: A 2D interacting-particle system governed by pairwise
forces.\\
\textbf{Query-Slots}: <$q_1$> <$q_2$> $\ldots$ <$q_{L_q}$>.
\end{tcolorbox}

\begin{tcolorbox}[promptbox, title={Spring-coupled particle prompt.}]
\small
\textbf{Task}: Given irregularly sampled observations of the particle system, predict each particle's future 2D position $(x,y)$ at the query slots.\\
\textbf{System}: A 2D particle system with spring-based interactions.\\
\textbf{Query-Slots}: <$q_1$> <$q_2$> $\ldots$ <$q_{L_q}$>.
\end{tcolorbox}

Each query-slot token
\texttt{<q}\textsubscript{$i$}\texttt{>}
corresponds to an irregular query timestamp $t_i^{(q)}$ and represents
a time-specific prediction request. For example, when predicting traffic flow at 10:01, 11:34, and 15:47, the
query slots
\texttt{<$q_1$>}, \texttt{<$q_2$>}, and \texttt{<$q_3$>}
are instantiated as
\texttt{<10:01>}, \texttt{<11:34>}, and \texttt{<15:47>}, respectively.
The temporal information is encoded using the same time-embedding
scheme introduced in Appendix~\ref{app:time-labeling}.




\subsection{Distribution Shift Analysis of NYC Datasets}
\label{app:distribution_shift_analysis}

\begin{figure}[t]
	\centering
	\renewcommand{\thesubfigure}{\normalfont\alph{subfigure}}
	\captionsetup[subfigure]{skip=0.1pt}
	\begin{subfigure}[t]{0.48\columnwidth}
		\centering
		\includegraphics[width=\textwidth,keepaspectratio]{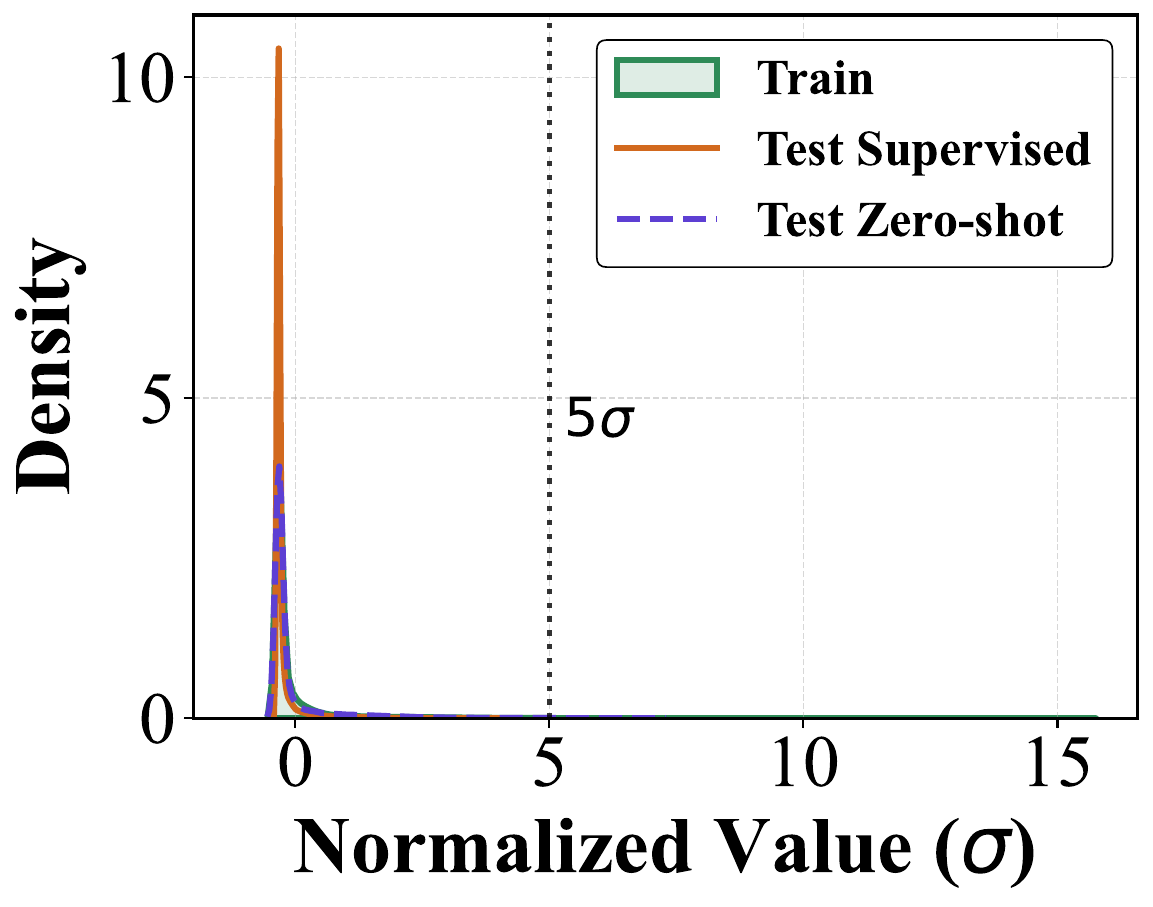}
		\caption{\normalfont Taxi: Inflow}
		\label{fig:dist-taxi-inflow}
	\end{subfigure}%
	\hfill
	\begin{subfigure}[t]{0.48\columnwidth}
		\centering
		\includegraphics[width=\textwidth,keepaspectratio]{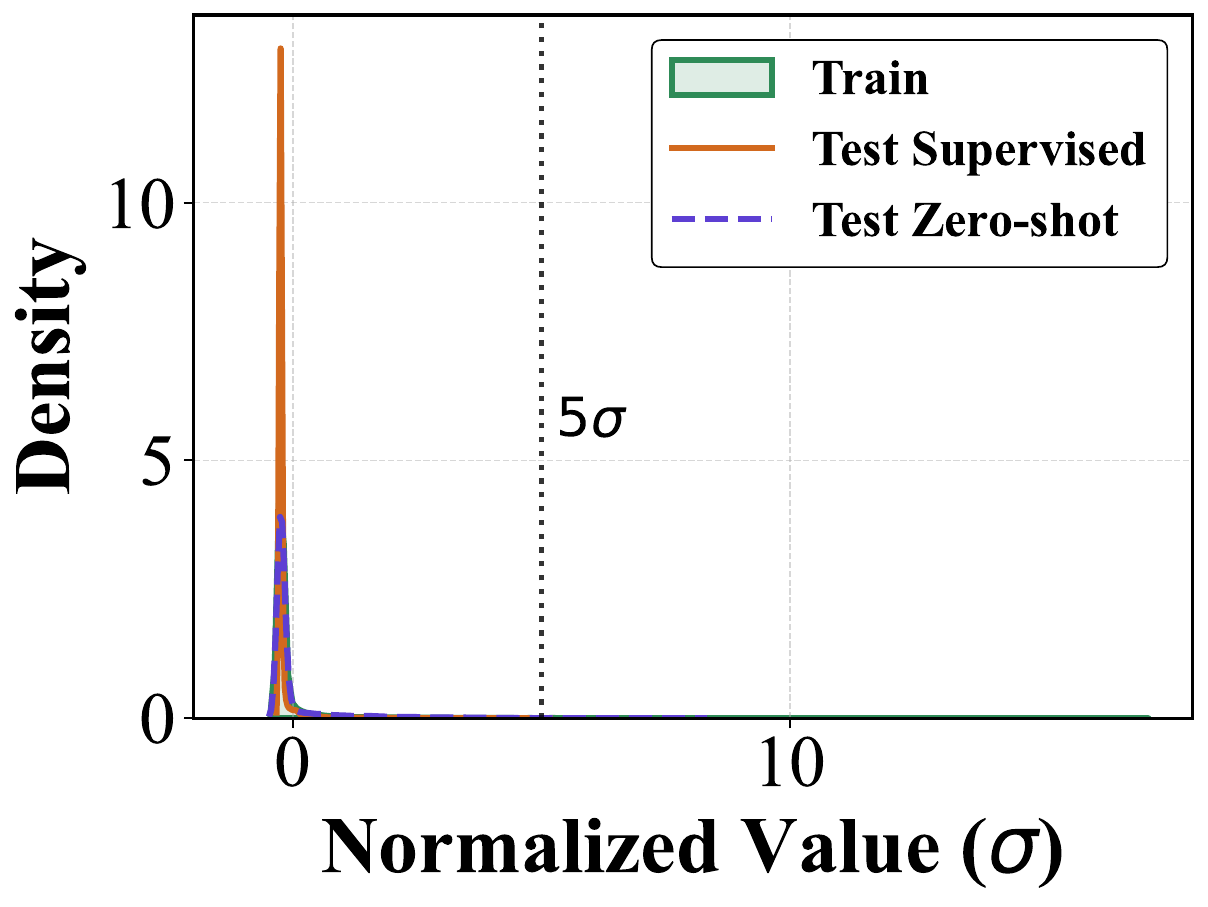}
		\caption{\normalfont Taxi: Outflow}
		\label{fig:dist-taxi-outflow}
	\end{subfigure}
	
	\vspace{0.5ex}
	
	\begin{subfigure}[t]{0.48\columnwidth}
		\centering
		\includegraphics[width=\textwidth,keepaspectratio]{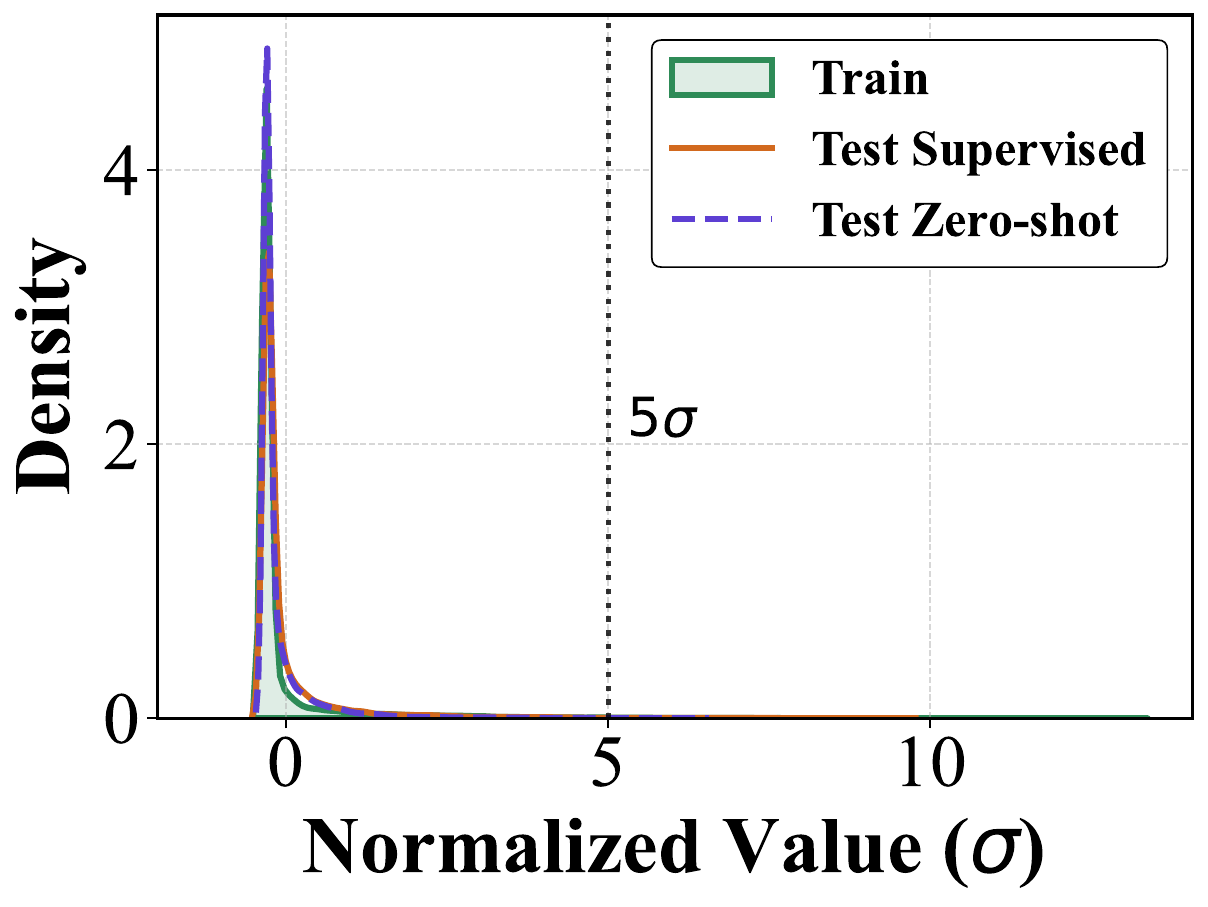}
		\caption{\normalfont Bike: Inflow}
		\label{fig:dist-bike-inflow}
	\end{subfigure}%
	\hfill
	\begin{subfigure}[t]{0.48\columnwidth}
		\centering
		\includegraphics[width=\textwidth,keepaspectratio]{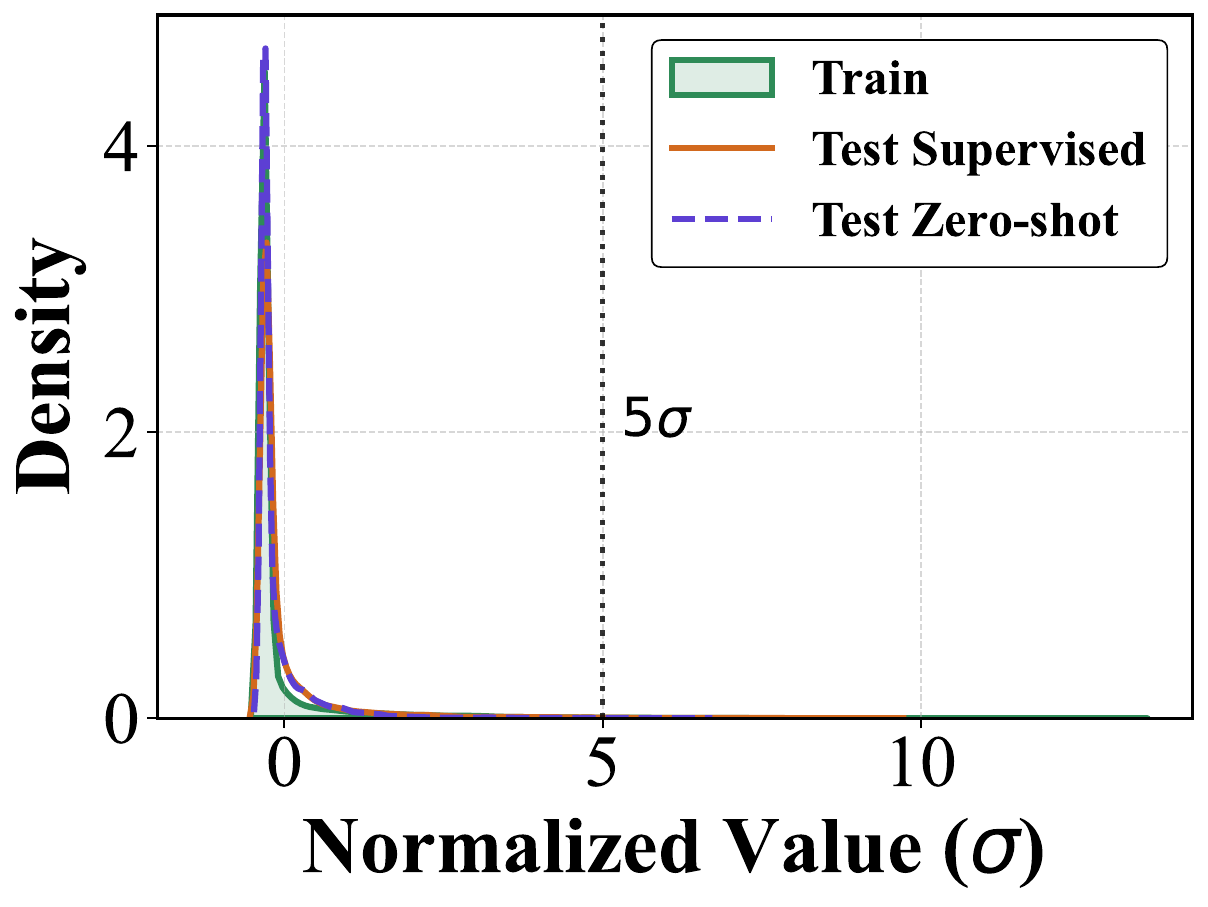}
		\caption{\normalfont Bike: Outflow}
		\label{fig:dist-bike-outflow}
	\end{subfigure}
	
	\vspace{0.5ex}
	
	\begin{subfigure}[t]{0.48\columnwidth}
		\centering
		\includegraphics[width=\textwidth,keepaspectratio]{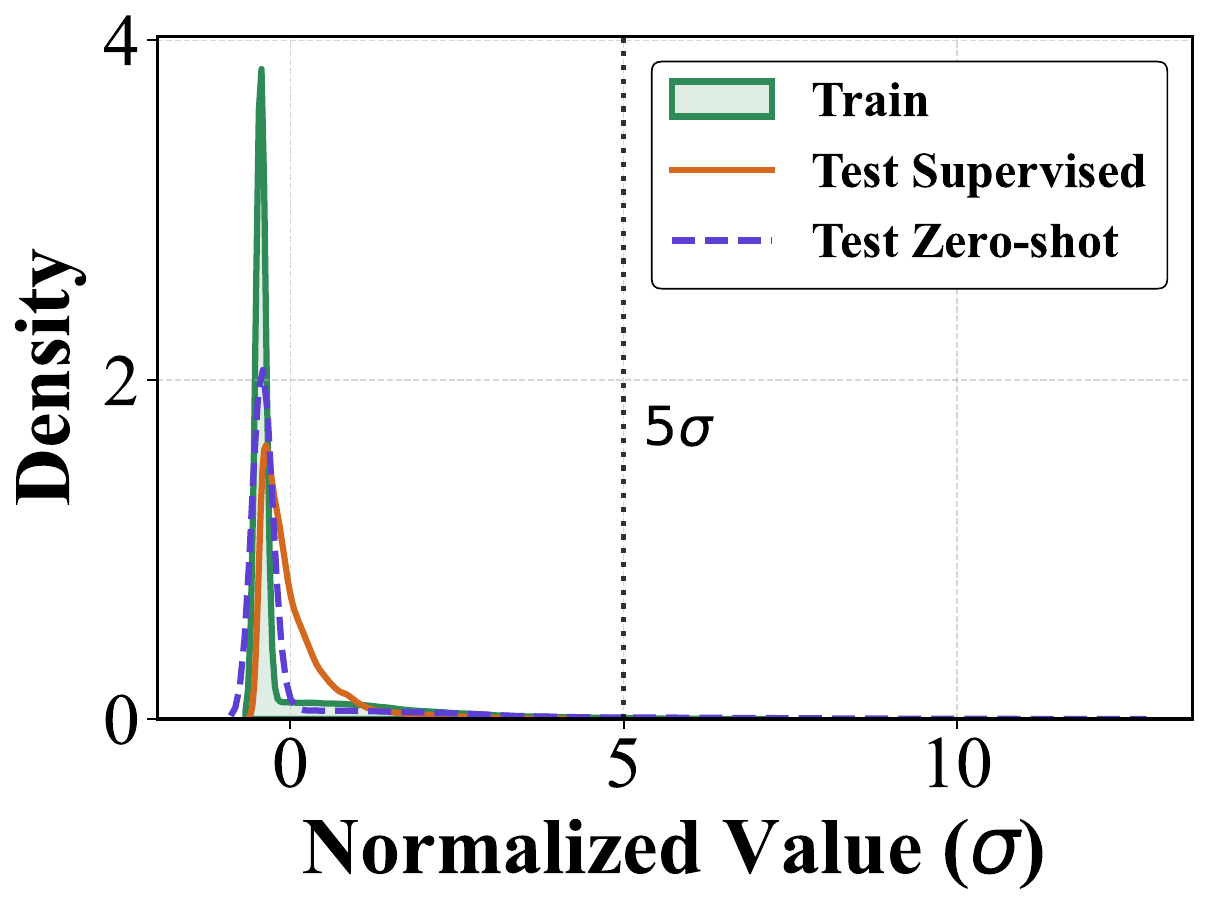}
		\caption{\normalfont Crime: robbery}
		\label{fig:dist-crime-robbery}
	\end{subfigure}%
	\hfill
	\begin{subfigure}[t]{0.48\columnwidth}
		\centering
		\includegraphics[width=\textwidth,keepaspectratio]{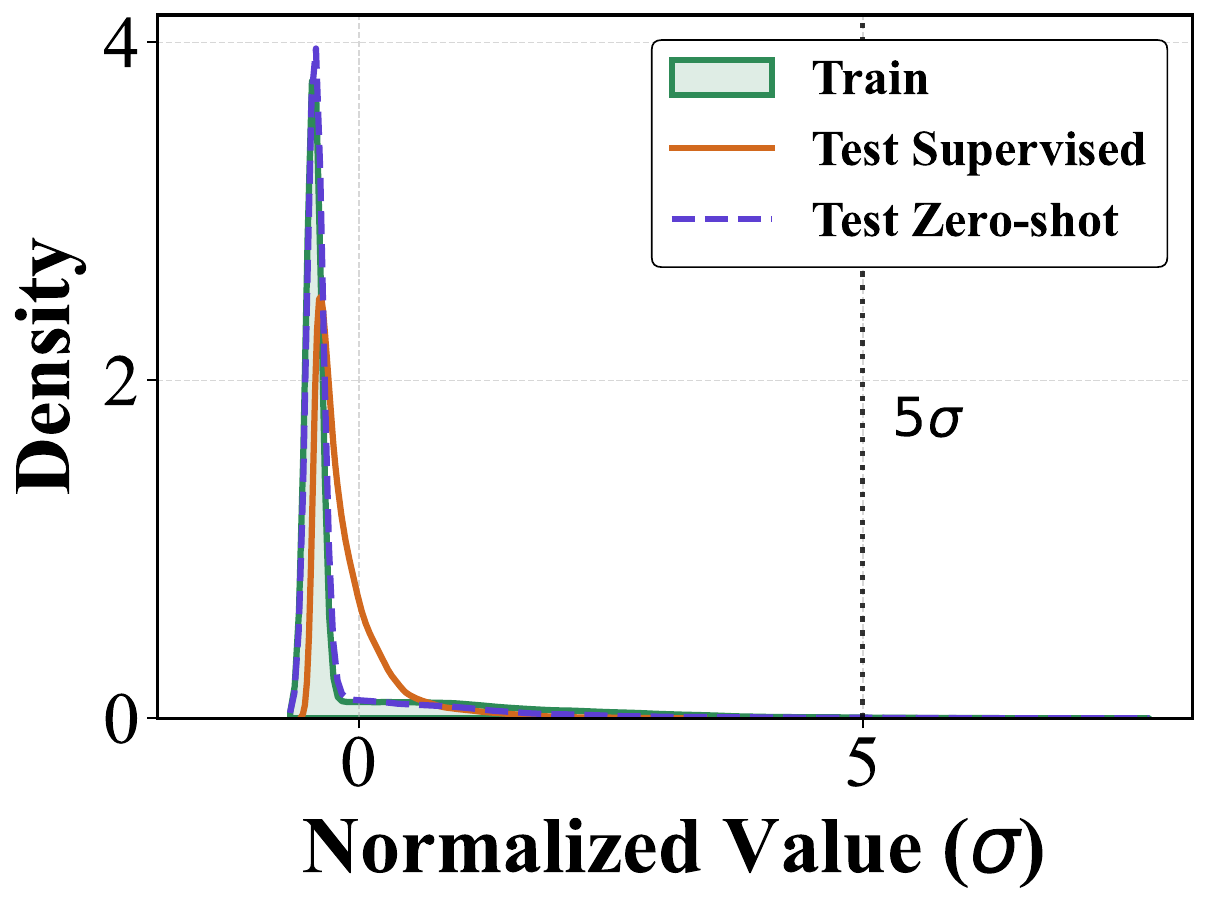}
		\caption{\normalfont Crime: burglary}
		\label{fig:dist-crime-burglary}
	\end{subfigure}
	
	\vspace{-2ex}
	\caption{Normalized distribution analysis across datasets. The vertical dashed line indicates the $5\sigma$ threshold used for data filtering.}
	\label{fig:dist}
	\vspace{-1ex}
\end{figure}


In this work, we use z-score normalized data for both training and testing on the NYC datasets, where normalization is performed using the mean and standard deviation computed from the training split. From Figure \ref{fig:dist}, we observe that all three datasets exhibit long-tail distributions. To avoid the impact of these long-tail distributions on model performance in irregular forecasting and ensure a fair comparison across methods, we remove points with values greater than $+ 5\sigma$, resulting in an approximately 2.2\% data removal rate. Importantly, this filtering is performed once during dataset preprocessing, and the resulting processed data are used identically for all baseline methods and LLMODE. Therefore, no method receives method-specific outlier treatment. 

Although the same preprocessing protocol is applied to all datasets and methods, differences remain between the training and test distributions.
We therefore further analyze two representative cases: the relatively large errors observed on NYCcrime and the comparatively small performance gains on NYCtaxi inflow.

\paragraph{Distribution Shift in NYCcrime.}
Despite this normalization and outlier removal, we find that the test distributions for the NYCcrime dataset still exhibit a mismatch compared to the training split. This discrepancy persists due to differences in the data distribution between the training and test sets. Since all models are trained and evaluated on the normalized data, this mismatch can disproportionately affect methods sensitive to scale or extreme values. In such cases, a small number of large errors can be amplified under squared-error metrics like MSE, leading to significantly worse performance on the NYCcrime dataset.

\paragraph{Distribution Shift in NYCtaxi Inflow.}
The distribution shift of NYCtaxi inflow is less visually apparent in Figure~\ref{fig:dist-taxi-inflow}; however, its summary statistics reveal a substantial contraction from the training split to the supervised test split.
After normalization using the training-split statistics, the standard deviation decreases from $1.000$ to $0.321$, while the 95th percentile decreases from $1.569$ to $0.195$.
These statistics indicate that the supervised test period has a considerably narrower and more stable value range than the training period.
Under such a concentrated evaluation distribution, dominant temporal patterns are relatively stable, and the advantage of explicitly reconstructing continuous-time dynamics may be less pronounced in terms of MAE.
As a result, interval-aware but token-discrete methods such as ISTS-PLM can remain competitive on this specific variable.
Nevertheless, LLMODE still achieves a slightly lower MSE on NYCtaxi inflow, suggesting better suppression of relatively large prediction errors.
Together with the stronger improvements observed on NYCbike, NYCcrime, and the physical-system benchmarks, we therefore regard the NYCtaxi inflow result as a dataset-specific case arising from its concentrated test distribution.

\subsection{Numerical Configuration Sensitivity}
\label{app:numerical_sensitivity}

LLMODE formulates the evolution between observations as differential equations and obtains continuous states through numerical integration. To verify that its performance is not tied to a particular numerical configuration, we further analyze the effects of the solver step size and solver choice on the Springs-X forecasting task.

\paragraph{Sensitivity to Solver Step Size.}
The default solver step size $\Delta t_0$ is set to the minimum sampling interval. As shown in Table~\ref{tab:step_size_sensitivity}, increasing the step size to $2\Delta t_0$ slightly increases the MAE and MSE from $0.0973/0.0191$ to $0.1039/0.0207$. This is expected because a finer integration grid provides a more accurate approximation of the continuous trajectory. Nevertheless, LLMODE with the larger step size still outperforms the best baseline, demonstrating its robustness to the choice of solver step size.

\begin{table}[t]
\centering
\small
\setlength{\tabcolsep}{8pt}
\caption{Sensitivity to the solver step size on the Springs dataset.}
\label{tab:step_size_sensitivity}
\begin{tabular}{lcc}
\toprule
Configuration & MAE & MSE \\
\midrule
Best baseline      & 0.1053 & 0.0237 \\
$\Delta t_0$       & \textbf{0.0973} & \textbf{0.0191} \\
$2\Delta t_0$      & 0.1039 & 0.0207 \\
\bottomrule
\end{tabular}
\end{table}

\paragraph{Sensitivity to ODE Solver.}
We further replace the default Euler solver with the higher-order RK4 solver. As reported in Table~\ref{tab:solver_choice}, RK4 provides no accuracy improvement but increases the ODE-solving latency from $0.5854$ to $1.6876$ ms per sample. This indicates that Euler already provides sufficient numerical accuracy when the step size matches the minimum sampling interval, while offering substantially lower computational cost. Together, these results show that LLMODE is not dependent on a specific solver or step-size configuration.

\begin{table}[t]
\centering
\small
\setlength{\tabcolsep}{5pt}
\caption{Sensitivity to the choice of ODE solver on the Springs dataset.}
\label{tab:solver_choice}
\begin{tabular}{lccc}
\toprule
Solver & MAE & MSE & Latency (ms/sample) \\
\midrule
Euler & \textbf{0.0973} & \textbf{0.0191} & \textbf{0.5854} \\
RK4   & 0.0986 & 0.0197 & 1.6876 \\
\bottomrule
\end{tabular}
\end{table}

\subsection{Detailed Token-Budget and Efficiency Results}
\label{app:token_budget}


\begin{table}[t]
\centering
\scriptsize
\caption{Token-budget and efficiency analysis on NYCbike zero-shot. Non-LLM forecasting methods are included for reference, while token-budget analysis focuses on LLM-based forecasting settings.}
\label{tab:token_budget}
\setlength{\tabcolsep}{10pt}
\begin{tabular}{@{}lcccc@{}}
\toprule
Variant & MAE & MSE & Latency (ms) & Tokens \\
\midrule

\multicolumn{5}{c}{\textit{LLM-based forecasting settings}}\\
\midrule

UrbanGPT 
& 0.0778 & 0.0292 & 646.73 & 665 \\

$K=1$
& 0.0863 & 0.0402 & 83.29 & 64 \\

$K=3$
& 0.0826 & 0.0301 & 83.49 & 68 \\

$K=5$
& \textbf{0.0745} & 0.0260 & 83.81 & 72 \\

$K=8$
& 0.0748 & \textbf{0.0254} & 84.10 & 78 \\

$K=30$
& 0.0782 & 0.0288 & 85.48 & 122 \\

$K=90$
& 0.0831 & 0.0322 & 86.04 & 242 \\

w/o compression
& 0.0839 & 0.0331 & 86.54 & 302 \\

\midrule
\multicolumn{5}{c}{\textit{Non-LLM forecasting methods}}\\
\midrule

MTGNN
& 0.0769 & 0.0357 & 9.87 & -- \\

tPatchGNN
& 0.0765 & 0.0323 & 10.63 & -- \\

\bottomrule
\end{tabular}
\end{table}

Table~\ref{tab:token_budget} provides the detailed numerical results corresponding to the token-budget efficiency analysis in Figure~\ref{fig:token_budget}, including token consumption and inference latency under different LLM-based forecasting settings. Compared with UrbanGPT, LLMODE reduces the inference latency from 646.73 ms to approximately 84 ms while substantially lowering the token budget, thereby reducing the computational overhead within the frozen-LLM forecasting paradigm. 

The reported token budget includes prompt tokens and both dynamic and context memory tokens. Unlike the appended spatio-temporal tokens used by UrbanGPT, the ODE-derived tokens are maintained as external memory and injected through gated cross-attention, without increasing the length
of the LLM self-attention sequence. Compared with full-trajectory exposure, fixed-budget compression reduces the reported token budget by approximately 75\% while maintaining or improving forecasting accuracy. Nevertheless, the compressed and full-trajectory settings
exhibit relatively similar inference latency, since changing the external-memory size has a more limited computational impact than changing the LLM input-sequence length.

Our objective is to explore a new modeling paradigm for irregular spatio-temporal forecasting by leveraging the representation capacity, prior knowledge, and transferability of pretrained LLMs. Accordingly, the efficiency analysis focuses on reducing the cost of processing external spatio-temporal evidence, rather than making an absolute speed comparison with all non-LLM methods. Nevertheless, we additionally report the inference latency of representative non-LLM forecasting methods as a computational reference. Their substantially lower latency reveals the remaining efficiency gap and motivates future work to further improve the computational efficiency of LLM-based forecasting while preserving the predictive and generalization benefits of pretrained language models.

\subsection{Additional Analysis of Gated Evidence Injection}
\label{app:injection_analysis}

\begin{figure*}[t]
	\centering
	
	\renewcommand{\thesubfigure}{\alph{subfigure}}
	\captionsetup[subfigure]{
		font=scriptsize,
		labelformat=parens,
		labelsep=space,
		justification=centering,
		singlelinecheck=true,
		skip=0.1pt
	}
	
	\includegraphics[width=0.4\textwidth,keepaspectratio]{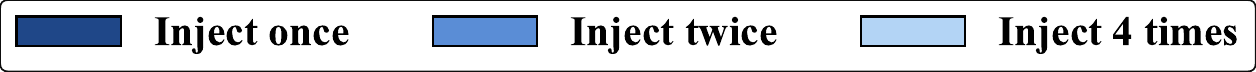}
	
	\vspace{0.2em}
	
	\begin{subfigure}[t]{0.14\textwidth}
		\centering
		\includegraphics[width=\textwidth,keepaspectratio]{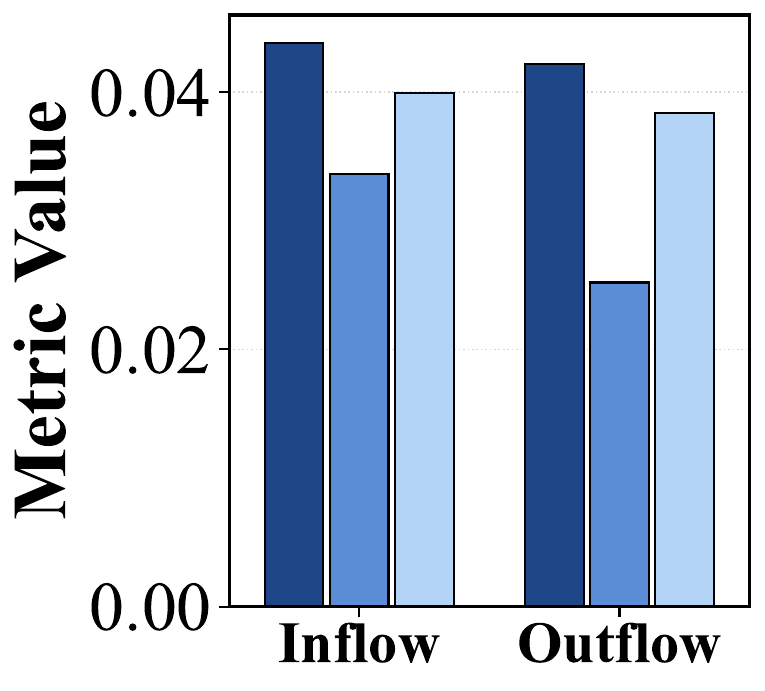}
		\caption{\mbox{\normalfont NYC Sup.\ MAE}}
		\label{fig:fig5-NYCsup-mae}
	\end{subfigure}%
	\hfill
	\begin{subfigure}[t]{0.14\textwidth}
		\centering
		\includegraphics[width=\textwidth,keepaspectratio]{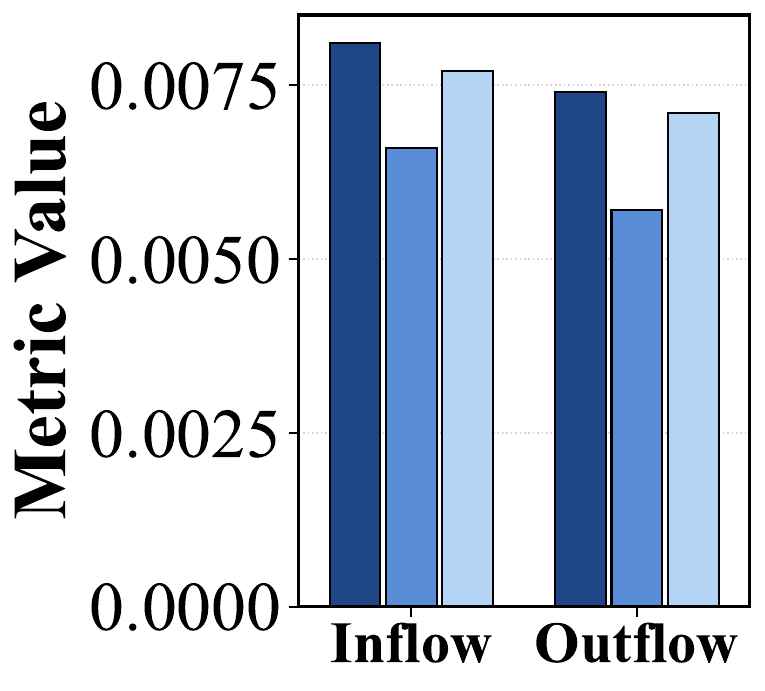}
		\caption{\mbox{\normalfont NYC Sup.\ MSE}}
		\label{fig:fig5-NYCsup-mse}
	\end{subfigure}%
	\hfill
	\begin{subfigure}[t]{0.14\textwidth}
		\centering
		\includegraphics[width=\textwidth,keepaspectratio]{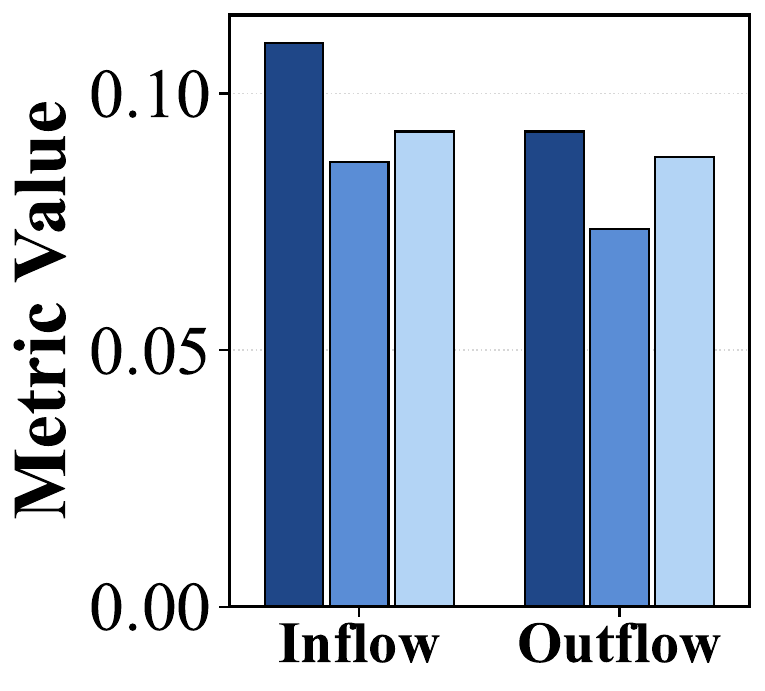}
		\caption{\mbox{\normalfont NYC Zero MAE}}
		\label{fig:fig5-NYCzero-mae}
	\end{subfigure}%
	\hfill
	\begin{subfigure}[t]{0.14\textwidth}
		\centering
		\includegraphics[width=\textwidth,keepaspectratio]{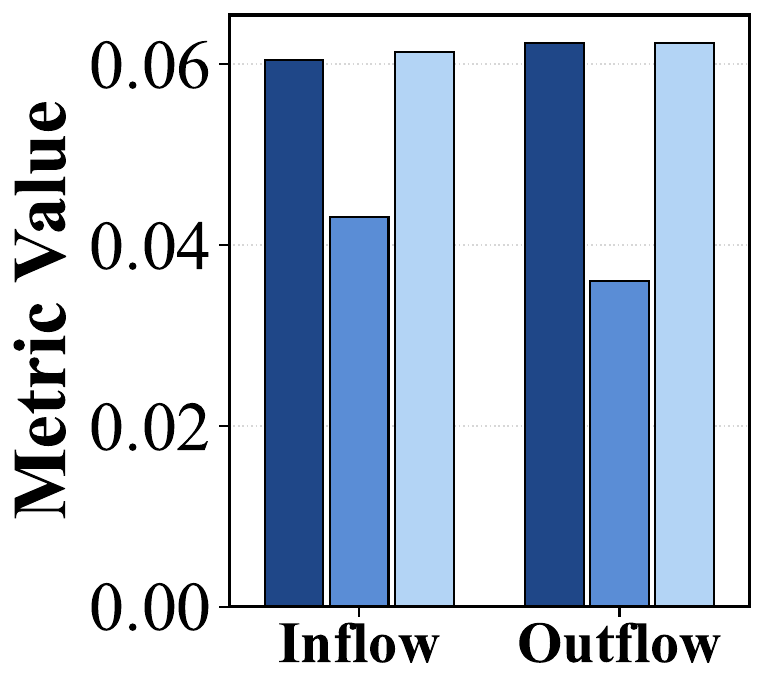}
		\caption{\mbox{\normalfont NYC Zero MSE}}
		\label{fig:fig5-NYCzero-mse}
	\end{subfigure}%
	\hfill
	\begin{subfigure}[t]{0.14\textwidth}
		\centering
		\includegraphics[width=\textwidth,keepaspectratio]{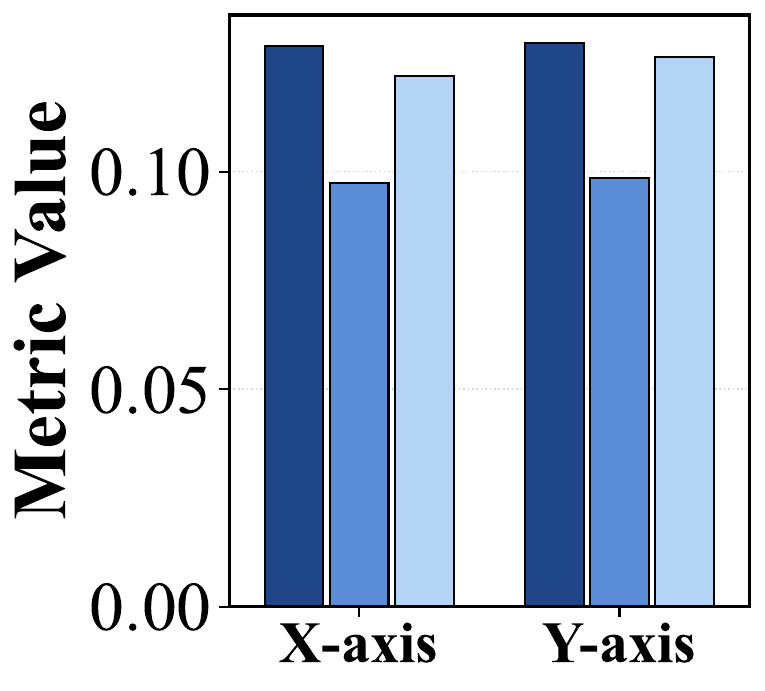}
		\caption{\mbox{\normalfont Springs MAE}}
		\label{fig:fig5-springs-mae}
	\end{subfigure}%
	\hfill
	\begin{subfigure}[t]{0.14\textwidth}
		\centering
		\includegraphics[width=\textwidth,keepaspectratio]{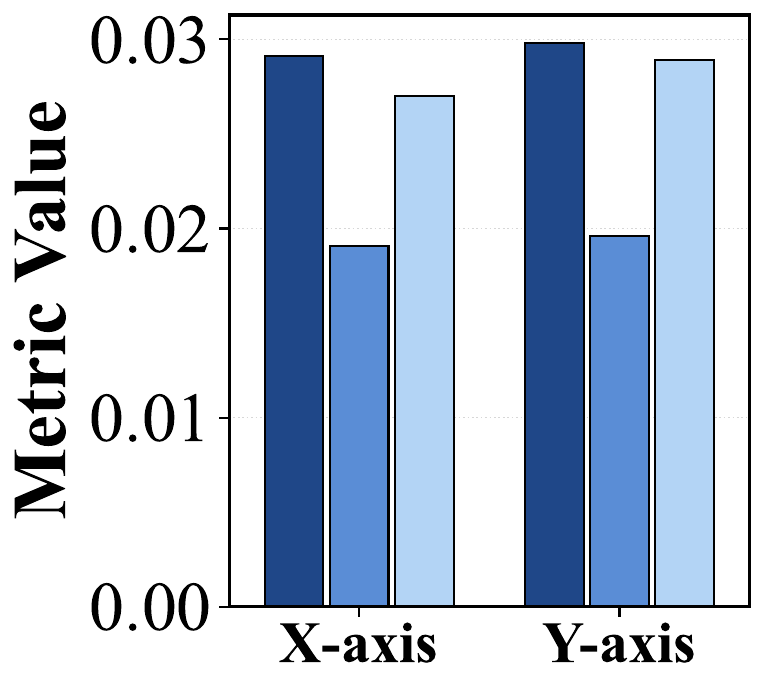}
		\caption{\mbox{\normalfont Springs MSE}}
		\label{fig:fig5-springs-mse}
	\end{subfigure}
	
	\vspace{-2ex}
	\caption{Performance under different injection frequencies. We use
	Vicuna-7B as the frozen backbone with 32 Transformer blocks and insert
	the dual-source gated injection modules between the blocks. We compare
	three configurations: one insertion after Block 1, two insertions with
	a 16-block interval, and four insertions with an 8-block interval.
	NYC Sup.\ and NYC Zero denote the supervised and zero-shot settings,
	respectively.}
	\label{fig:injection-frequencies}
	\vspace{-1ex}
\end{figure*}

\paragraph{Insertion Frequency and Number.}
The frozen Vicuna-7B backbone contains 32 Transformer blocks.
We insert the dual-source gated injection modules between these blocks and jointly vary their insertion frequency and number.
Specifically, we consider three configurations: 
(i) one early injection at Block~1,
(ii) two injections with an interval of 16 blocks, and
(iii) four injections with an interval of 8 blocks.
Figure~\ref{fig:injection-frequencies} reports the results on NYCtaxi-Supervised, NYCtaxi-Zero-shot, and Springs.

As shown in Figure~\ref{fig:injection-frequencies},  inserting two modules every 16 blocks achieves the best overall performance and consistently improves over using only one early injection.
Increasing the insertion frequency to every 8 blocks does not yield further gains and can even degrade accuracy.
This suggests that overly frequent injection may introduce redundant updates into the frozen representations.
Moreover, each additional module requires extra cross-attention computation at another depth of the LLM, increasing both computational cost and inference latency.
We therefore adopt two gated injection modules with a 16-block interval as the default configuration.

\paragraph{Gate Evolution and Effective Contribution.}
\label{sec:contrib_details}
Under the default every-16-blocks configuration, two gated injection modules are inserted into the frozen Vicuna-7B backbone.
Layer~0 and Layer~1 in Figure~\ref{fig:app_gate_evolution} denote the first and second injection modules, respectively.
For each module, the learnable gate parameter $g$ is transformed into the effective gate gain
$\alpha=\tanh(g)$.
The solid curves in Figure~\ref{fig:app_gate_evolution} show the evolution of $\alpha$ throughout training.

Because the gate gain alone does not fully characterize the actual magnitude of the injected update, we additionally report the normalized Effective Contribution:
\begin{equation}
\mathcal{C}
=
\frac{\alpha \lVert h \rVert_2}
     {\lVert h_{\mathrm{prompt}} \rVert_2},
\end{equation}
where $h$ denotes the representation produced by the corresponding cross-attention branch, and $h_{\mathrm{prompt}}$ denotes the prompt-token representation before the gated residual update.
The metric therefore measures the magnitude of the injected update relative to the current backbone representation.
The shaded regions in Figure~\ref{fig:app_gate_evolution} visualize this normalized contribution.

Starting from zero initialization, the gate gains of both modules gradually increase and eventually stabilize.
This behavior shows that the model progressively opens the injection paths, incorporates the external dynamic and contextual tokens, and finally converges to stable injection strengths.
Although Layer~0 generally learns a smaller gate gain than Layer~1, the gate values cannot be compared in isolation because the actual update also depends on the magnitude of the attended representation $h$.
The Effective Contribution provides this complementary view and confirms that both injection modules maintain meaningful non-zero updates after convergence.

\subsection{Qualitative Case Study}
\label{app:case_study}

Figure~\ref{fig:nyc-taxi-case-study} presents a representative
forecasting example on NYCtaxi under the supervised setting, comparing
LLMODE with the strong irregular-aware baseline
tPatchGNN~\cite{tPatchGNN}.
For each evaluation window, the models use a 24-hour history to predict
the following 12 hours.
For visualization, forecasts from consecutive windows are arranged
chronologically to form a 124-hour trajectory.

Across the evaluation period, LLMODE follows the ground-truth trends
more consistently and exhibits smaller deviations than tPatchGNN.
The enlarged region around hours 58--64 further illustrates their
difference during a sharp demand variation.
LLMODE more accurately captures both the magnitude and temporal
location of the peak, whereas tPatchGNN exhibits larger amplitude
errors and temporal misalignment.
Similar behavior is observed for both inflow and outflow, indicating
that the improvement is not limited to a single forecasting variable.
Overall, the case study qualitatively demonstrates that LLMODE better
preserves complex temporal patterns and produces more stable forecasts
under irregular spatio-temporal observations.

\subsection{Baseline Details}
\label{app:baselines}
To ensure fair comparisons across methods with different temporal assumptions, we apply a unified adaptation for baselines that only operate on a \emph{regular} time grid. On the input side, we linearly interpolate the irregular observations along time to obtain regularly sampled sequences before feeding them into such models. On the evaluation side, the protocol differs across datasets. For the NYC datasets, the observations are event-driven and do not provide an underlying complete regularly sampled series from which a missingness mask can be defined, so we standardize the target as predicting the next 12 \emph{regular} time steps and evaluate all methods on these aligned horizons. For the physics simulation datasets, the original trajectories are generated and recorded on a fixed-step grid, and we create irregular inputs by subsampling this complete regular sequence and specifying irregular query times as prediction targets. This construction provides explicit indices indicating which future regular steps correspond to the query times. Therefore, for methods that can only produce regular-step forecasts, we ask them to predict the next 60 regular steps and then select the predictions at the indexed query-aligned steps for evaluation, enabling consistent assessment under irregular query horizons.

We compare our method with a diverse set of baselines, including pre-trained language model based approaches and graph neural network based methods, covering both irregularly sampled and regularly sampled spatio-temporal forecasting settings.

\textbf{TGCN} \cite{TGCN} is a temporal graph convolutional network for spatio-temporal forecasting, which combines graph convolution to model spatial dependencies with recurrent units to capture temporal dynamics, and is originally designed for regularly sampled data defined over a fixed graph.

\textbf{STSGCN} \cite{STSGCN} is a spatial-temporal graph convolutional network for spatio-temporal forecasting, which jointly models spatial and temporal dependencies through localized spatial-temporal subgraphs. It is originally designed for regularly sampled spatio-temporal data defined over a fixed graph.

\textbf{MTGNN} \cite{MTGNN} utilizes a learnable graph structure to model multivariate temporal correlations. MTGNN employs 1-D dilation convolutions to generate temporal representations.
  
\textbf{BiTGraph} \cite{BiTGraph} is a biased temporal convolution based graph network for time series forecasting with missing values, which combines graph neural networks to model spatial dependencies with biased temporal convolutions to handle irregular sampling patterns and missing observations.

\textbf{FourierGNN} \cite{FourierGNN} transforms multivariate sequences into the frequency domain to model spatio-temporal dependencies via spectral graph convolutions. 

\textbf{ISTS-PLM} \cite{ISTS-PLM} is a pre-trained language model based approach for irregularly sampled time series forecasting, designed to handle non-uniform time gaps and missing observations through LLM-compatible series representations.

\textbf{UrbanGPT} \cite{li2024urbangpt} integrates a spatio-temporal dependency encoder with a Large Language Model to capture complex urban dynamics via spatio-temporal instruction tuning. 

\textbf{GPT4TS} \cite{GPT4TS} reframes time series forecasting as a language modeling task by processing patch-based representations through a frozen pre-trained GPT-2 backbone. 

\textbf{GRU-ODE} \cite{GRU-ODE} combines Gated Recurrent Units with Neural Ordinary Differential Equations to model continuous-time latent states for irregularly sampled time series.

\textbf{tPatchGNN} \cite{tPatchGNN} is a graph neural network based approach for modeling irregularly sampled time series, which represents observations as temporal patches and captures temporal dependencies through patch-level message passing to handle non-uniform time gaps and missing observations.

\textbf{ViTST} \cite{ViTST} is a vision transformer based model for irregularly sampled time series analysis, which transforms irregular multivariate time series into line graph images and leverages vision transformers for representation learning and forecasting.

\end{document}